\documentclass{article}

\usepackage[preprint]{neurips_2026}

\usepackage[utf8]{inputenc} 
\usepackage[T1]{fontenc}    
\usepackage{hyperref}       
\usepackage{url}            
\usepackage{booktabs}       
\usepackage{amsfonts}       
\usepackage{nicefrac}       
\usepackage{microtype}      
\usepackage{xcolor}         

\usepackage{amsmath}
\usepackage{amssymb}
\usepackage{mathtools}
\usepackage{amsthm}

\usepackage{algorithm}
\usepackage{arydshln}
\usepackage{multirow}
\usepackage{placeins}
\usepackage{algorithmic}
\usepackage{graphicx} 

\usepackage{wrapfig}
\usepackage{caption}
\title{Minimizing the Hidden Cost of Scales: Graph-Guided Ultra-Low-Bit Quantization for Large Language Models}

\author{%
  Rayyan Abdalla \\
  \And
  Amir Hussein \\
  \And
  Min Wu
  \And
  Dinesh Manocha
}

\begin{document}

\maketitle

\begin{abstract}
Post-training quantization (PTQ) is critical for the efficient deployment of large language models (LLMs). Recent ultra-low-bit PTQ methods rely on rigid weight-saliency assumptions or position heuristics, introducing substantial hidden scaling overhead.
We propose \textsc{SAGE-PTQ} (\textbf{S}aliency-\textbf{A}ware \textbf{G}raph-guided \textbf{E}fficient PTQ), a novel ultra-low-bit quantization framework for LLMs that minimizes hidden scaling cost. \textsc{SAGE-PTQ} separates salient and unsalient weights using distributional statistics, then models subsampled unsalient weights as a sparse graph to estimate the optimal number of groups per layer. SAGE-PTQ applies dual-mode quantization, assigning multi-bit precision to salient weights and binarizing unsalient weights. To reduce scaling overhead, SAGE-PTQ uses one per-channel scale for salient weights and one scalar per unsalient group. Finally, SAGE-PTQ implements adaptive saliency thresholding to select the optimal saliency ratio per matrix.
\textsc{SAGE-PTQ} achieves $1.03$ weight bits and only $0.004$ scaling bits per matrix on average, outperforming state-of-the-art methods such as BiLLM and PB-LLM. On LLaMA-3-8B, \textsc{SAGE-PTQ} achieves $6.74$ WikiText2 perplexity, compared to $55.8$ for BiLLM, while using less than $50\%$ of BiLLM's GPU memory. On LLaMA-2-70B, \textsc{SAGE-PTQ} provides $1.5\times$ faster decoding on one NVIDIA L40 GPU, demonstrating practical inference efficiency.

\end{abstract}
\vspace{-0.4cm}
\section{Introduction}
\vspace{-0.2cm}
Large language models (LLMs), including OPT~\cite{zhang2022opt}, LLaMA~\cite{touvron2023llama, touvron2023llama2,grattafiori2024llama3}, and DeepSeek~\cite{deepseek2024}, achieve strong language and reasoning performance~\cite{chen2021codex, wei2022chain, openai2023gpt4}. 
However, their size creates substantial memory demands,(e.g: LLaMA-2-70B requiring ~140~GB in FP16), limiting deployment. 
Moreover, auto-regressive decoding is dominated by GEMV operations~\cite{zeng2025abq}, causing  higher inference latency.

To mitigate memory and latency bottlenecks, compression techniques such as quantization~\cite{liu2022post, jiao2019tinybert}, pruning~\cite{frantar2023gptq, ma2023dynamic}, and distillation~\cite{gou2021knowledge, tunstall2023hf} have been explored. 
Among these, quantization is particularly attractive for efficient deployment without retraining. 
PTQ methods perform well at 4-8 bit precision~\cite{dettmers2022llm, xiao2023smoothquant, frantar2022gptq}, while recent work has pushed into lower-bit regimes. 
AWQ~\cite{lin2024awq} and GPTQ~\cite{frantar2022gptq} extend PTQ to 3-bit precision but degrade noticeably at lower bit widths.

Binarized PTQ methods, such as PB-LLM~\cite{shang2023pb} and BiLLM~\cite{huang2024billm}, 
treat quantization-sensitive outliers as \textit{\textbf{salient}} weights requiring higher precision, and treat quantization-insensitive inliers as\textit{ \textbf{unsalient}} weights that are binarized. 
However, they rely on rigid grouping and heuristic thresholds, which fail to generalize across models. 
We analyze prior limitations by studying weight statistics and saliency patterns across diverse LLM families (Appendix~\ref{stat_study}) and identify four core issues in PTQ:  
\textit{(1) Saliency handling}: salient weights vary across layers and require adaptive high-precision treatment. 
GPTQ~\cite{frantar2022gptq} poorly handles outliers due to calibration dependence, 
while PB-LLM~\cite{shang2023pb} applies fixed saliency thresholds across layers.  
\textit{(2) Weight partitioning}: Rigid or uniform partitioning fails on long-tailed distributions. 
BiLLM~\cite{huang2024billm} assumes Gaussian weights, limiting generalization across architectures. 
\textit{(3) Suboptimal quantization}: Most methods rely on simple min--max scaling, causing large quantization error. 
Sub-2-bit methods such as GPTQ, PB-LLM, and BiLLM further assume block-based scaling, introducing hidden overheads 
that can exceed 1 bit per weight and offset binarization gains.  
\textit{(4) Lookup efficiency}: Efficient group index restoration is critical for deployment, 
yet BiLLM~\cite{huang2024billm} and PB-LLM~\cite{shang2023pb} ignore runtime-aware decoding and lookup, limiting deployment.

\begin{figure*}[!t]
    \vspace{-0.2cm}
    \includegraphics[
        width=14.8cm,
        height=4.6cm,
    ]{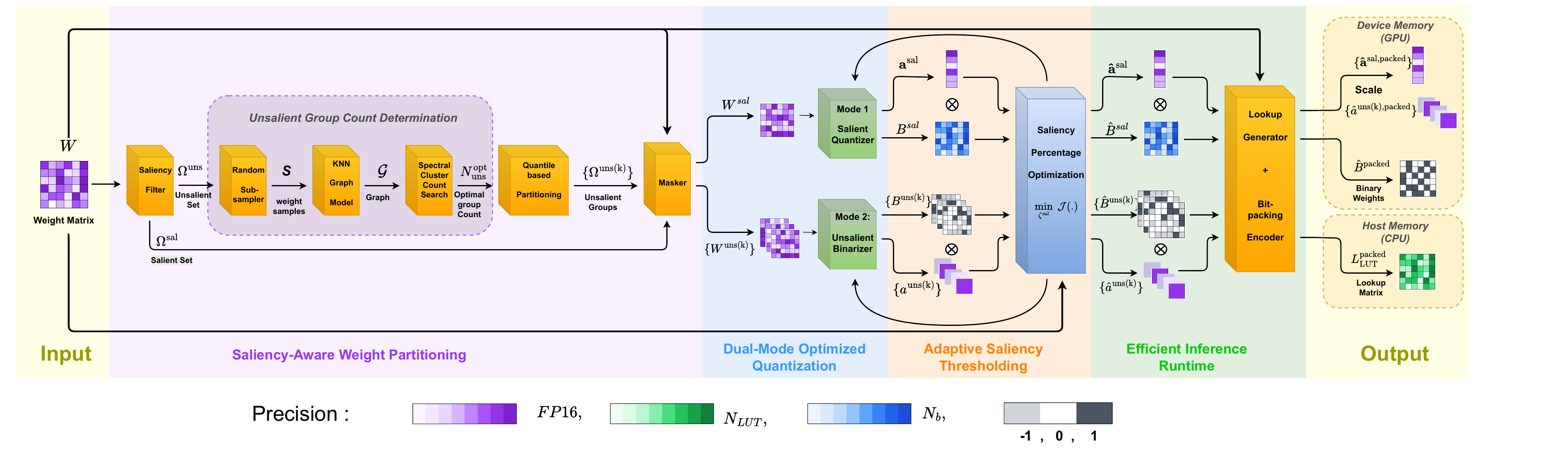}
    \caption{Illustration of the SAGE-PTQ quantization pipeline. The framework input is the weight matrix and outputs quantized matrix with minimal scale overhead and minimal lookup metadata}
    \label{fig1}
    \vspace{-0.5cm}
\end{figure*}

\noindent {\bf Contributions:} We target the core limitations of binarized PTQ methods by proposing \textbf{SAGE-PTQ}, a novel framework for optimized quantization of LLMs. Our approach consists of the following:
\begin{itemize}
\vspace{-0.15cm}
\item We propose a novel \textbf{weight partitioning} approach that separates salient and unsalient weights per matrix, using criteria based on weight distribution. We also introduce a graph-guided algorithm that adaptively estimates the optimal number of unsalient weight groups.
\vspace{-0.3cm}
\item We design an \textbf{optimized dual-mode quantizer}. 
Our approach assigns arbitrary-bit precision to salient weights and binarizes unsalient weights, while minimizing scaling overhead using a single scalar per unsalient group and a single channel-wise scale per salient group. 
With a 4-bit saliency budget, our quantizer achieves an average of $1.03$ bits per weight and only $0.004$ scaling bits, outperforming BiLLM with over $5\times$ lower perplexity on LLaMA models.
\vspace{-0.3cm}
\item We formulate an optimization objective for \textbf{adaptive saliency thresholding} to determine the optimal saliency ratio per matrix under quantization-error and storage constraints. On LLaMA-7B, this yields $\sim 50\%$ lower perplexity with $\sim 30\%$ fewer salient-weight bits.
\end{itemize}
\vspace{-0.2cm}
The SAGE-PTQ pipeline is illustrated in Figure~\ref{fig1}. SAGE-PTQ improves deployment practicality by introducing an efficient group lookup mechanism, reducing lookup costs by 25\% compared to BiLLM and achieving a 1.5$\times$ speedup on 70B models deployed on a single GPU. SAGE-PTQ is a model-agnostic framework, validated across LLaMA, OPT, Vicuna, DeepSeek and Qwen models ranging from 1.3B to 70B parameters, where it consistently achieves state-of-the-art (SoTA) performance.
\vspace{-0.25cm}
\section{Related Work}
\vspace{-0.2cm}
\subsection{LLM Post-Training Quantization}
\vspace{-0.2cm}
Post-training quantization (PTQ) enables efficient LLM deployment by compressing models into low-bit formats without retraining. Early PTQ methods targeted weight-only quantization using rounding optimization or compensation~\cite{frantar2022gptq,nagel2020up, li2021brecq}, while later approaches extended to activations via scaling and incoherence modeling~\cite{xiao2023smoothquant,yao2022zeroquant, tseng2024quip}. Recent efforts introduced structural priors and learned quantization~\cite{kalyvianaki2011sqpr, kim2023squeezellm, shao2023omniquant} and block-clustered approaches \cite{elangovan2025bcq}, yet binarization remains challenging. Hybrid methods~\cite{shang2023pb, huang2024billm} retain salient weights in high precision but rely on heuristics and incur overhead. Our method, SAGE-PTQ, addresses these limitations by optimizing weight partitioning and saliency allocation, achieving SoTA binarized PTQ results across LLM families.
\vspace{-0.2cm}
\subsection{Saliency-Aware Precision Allocation}
\vspace{-0.2cm}
Saliency-aware quantization improves model fidelity by assigning higher precision to salient weights identified via metrics like magnitude, Hessian sensitivity, or reconstruction error. HAWQ~\cite{dong2019hawq} uses Hessian trace for layer-wise bitwidths, while QDrop~\cite{wei2022qdrop} applies stochastic dropout during calibration. For LLMs, AWQ~\cite{lin2024awq} retains high-activation channels in FP16; SmoothQuant~\cite{xiao2023smoothquant} rescales outliers; PB-LLM~\cite{shang2023pb} binarizes inliers and preserves outlier weights in INT8; BiLLM~\cite{huang2024billm} searches salient columns using residual error. SAGE-PTQ advances this line by combining element-wise saliency filtering with adaptive thresholding to optimally allocate saliency bits under memory constraints.
\vspace{-0.3cm}
\subsection{Graph-guided Partitioning}
\vspace{-0.2cm}
Graph-based methods such as KNN graphs model local affinities and are widely used in high-dimensional data analysis~\cite{zelnik2005selftuning,ng2002spectral}. Spectral clustering~\cite{luxburg2007tutorial} leverages the graph Laplacian to identify coherent groups, with the Silhouette Score~\cite{rousseeuw1987silhouettes} selecting the optimal cluster count. Prior quantization methods, like vector quantization~\cite{gong2014vectorquant} and CLIP-Q~\cite{tung2018clipq}, applied clustering without graph structures. SAGE-PTQ introduces graph-guided modeling for unsalient weights via sparse KNN, applying spectral clustering and Silhouette Score to enable accurate, optimal weight group count discovery.
\vspace{-0.25cm}
\section{SAGE-PTQ Method}
\vspace{-0.21cm}
\paragraph{Preliminaries:}
We employ a generalized formulation that supports both binarized and multi-bit precision for weight quantization. Our formulation covers both structured (channel-wise) and unstructured (element-wise) weight partitioning. Given a pretrained weight matrix \( W \in \mathbb{R}^{m \times n} \), we formulate the problem of finding an approximation \( \hat{W} \) as minimizing the reconstruction error: \vspace{-0.1cm}
{\small
\begin{equation} \label{opt_eq}
\min_{\hat{W}} \|W - \hat{W}\|^2_F, \quad \text{where} \quad \hat{W} = A B.
\end{equation}
}
where \( A \in \mathbb{R}^{m \times m} \) is a diagonal scaling matrix, 
\( B \in \mathbb{R}^{m \times n} \) contains quantized entries, where each
\( b_{ij} \in [-1, 1] \) is represented using \( N_b \)-bit encoding.
Equation~\eqref{opt_eq} uses the Frobenius norm \( \|\cdot\|_F \) to minimize the residual energy between \( W \) and \( \hat{W} \). This approximation is particularly effective when salient weights (outliers) form a small fraction of total weights, as described in Appendix \ref{stat_study}.
\vspace{-0.21cm}
\subsection{Saliency-Aware Weight Partitioning}
\vspace{-0.15cm}
\subsubsection{Salient-Weight Filtering}
\vspace{-0.15cm}
Salient weights are quantization-sensitive and thus require higher precision to preserve model accuracy.
Saliency selection can use magnitude metric or Hessian(loss) metric. We evaluate both saliency metrics and analyze their effects on quantized model performance and scaling overhead.\\
Consistent with the findings of~\cite{shang2023pb}, our ablation study in Appendix~\ref{sal_ablation} shows that magnitude saliency more reliably identifies salient weights than Hessian (loss) measures, leading to lower quantization error with minimal scaling overhead.
Our analysis further shows that Hessian-based methods~\cite{frantar2023gptq,huang2024billm} are effective when combined with position-based partitioning. This subdivides weights into many local blocks and enables Hessian information to correct localized quantization errors. However, aggressive partitioning introduces large and untracked scaling overhead.
To explain this behavior, we analyze weight distributions across diverse model layers. We find that most layers exhibit approximately Gaussian distributions, while early layers deviate significantly from normal trend due to the presence of heavy-tailed outliers, as shown in Appendix~\ref{stat_study}.\\
Building on~\cite{wang2025bi}, our salient weight filtering method adaptively selects a saliency threshold \( \zeta_l^{\text{sal}} \) for each weight matrix \( l \in \{1,\dots,N_l\} \).
This threshold is constrained by a heuristic upper bound \( \zeta_l^{\text{sal,max}} \), which is determined by pre-set saliency storage budget.
We further analyze the choice of the saliency upper bound in Appendix~\ref{sensi_param}.\\
For each layer  {\small \( l \)}, we partition the set of weight values {\small \( \Omega_l \)} into salient and non-salient subsets, such that:
{\small \( \Omega_l = \Omega_l^{\text{sal}} \cup \Omega_l^{\text{uns}} \)}, with
{\small \( \Omega_l^{\text{sal}} \cap \Omega_l^{\text{uns}} = \varnothing \)}.
Let {\small \( \beta_l \)} and {\small \( \gamma_l \)}, denote the arithmetic mean and standard deviation of {\small \( \Omega_l \)}, respectively,
we define the salient subset:
{\small 
\begin{equation}
    \Omega_l^{\text{sal}} = \{ w \in \Omega_l \mid |w| > \beta_l + \gamma_l \, \Phi^{-1}(\zeta_l^{\text{sal}}/2) \}
\end{equation}}
where {\small \( \Phi^{-1}(\cdot) \)} is the probit function.
This formulation is motivated by empirical analysis of weight statistics (Appendix~\ref{stat_study}), which shows that weight distributions are nearly symmetric around {\small \( \beta_l \)}.
\vspace{-0.2cm}
\subsubsection{Graph-Guided Estimation of Unsalient Group Count}
\vspace{-0.1cm}
Unsalient weights comprise the majority of the weight matrix and are therefore targeted for aggressive quantization. However, these weights often exhibit non-uniform magnitude distributions, which limits the effectiveness of uniform min–max scaling schemes.\\
Prior work~\cite{frantar2023gptq,shang2023pb,huang2024billm} partitions unsalient weights using positional blocks without accounting for weight distributions, requiring channel-wise quantization for each group and introducing untracked scaling overhead.
While quantile-based partitioning suits Gaussian-like distributions, our empirical analysis in Appendix~\ref{stat_study} shows that unsalient weight distributions vary considerably across layers, especially early layers that deviate from normality, limiting the applicability of a fixed partitioning strategy.
Using a fixed number of quantiles therefore fails to capture layer-wise weight heterogeneity unless combined with fine-grained partitioning, which introduces substantial scaling cost.
Clustering methods such as K-means and spectral clustering provide alternative partitioning strategies but are impractical for LLM weight partitioning~\cite{zhang2025towards}.
For large unsalient sets \( \Omega_l^{\text{uns}} \), clustering methods are sensitive to initialization and prune to cluster misassignments when weight variance is low. \\
To address the limitations of unsalient weight partitioning, we propose a graph-guided algorithm that estimates the optimal number of unsalient groups \(N_{\text{uns},l}^{\text{optimal}}\) for each matrix \(l\), reducing scaling cost and lookup overhead. 
The algorithm models unsalient weights as a graph \(\mathcal{G}\), where each node represents a weight and each edge encodes pairwise affinity based on Euclidean distance. 
Because LLM weight matrices are large, we apply randomized subsampling to select \(S_l\) representative weights from the unsalient set \(\Omega_l^{\text{uns}}\) for efficient graph construction.\\
The sampling focuses on columns that capture attention patterns.
To further improve efficiency, we construct a sparse \(K\)-nearest-neighbor (KNN) graph as the affinity matrix, preserving the local structure of the unsalient weight distribution. 
Details on graph construction cost and hyperparameter selection are provided in Appendix~\ref{hyper_param}.
Spectral clustering is then applied to the affinity matrix to evaluate candidate unsalient group counts. 
The candidate range is constrained by the host memory budget to ensure that the average lookup bitwidth remains within \(N_{\text{LUT}}\).
We use the Silhouette Score \(s_r\) to evaluate each candidate group count, defined as
{\small
\begin{equation} \label{s_r}
s_r = \frac{1}{|S_l|} \sum_{i=1}^{|S_l|} \frac{v(i) - u(i)}{\max\{u(i), v(i)\}},
\end{equation}}
\vspace{0.15cm}
where \(|S_l|\) is the size of the subsampled unsalient set, \(r\) is the candidate group count, \(u(i)\) is the average intra-group distance, and \(v(i)\) is the minimum average distance from weight \(i\) to weights in other groups. 
The Silhouette Score captures whether each weight is closer to its assigned group than to other groups. 
The optimal group count, \(N_{\text{uns},l}^{\text{optimal}}\), maximizes \(s_r\).
Algorithm~\ref{alg1} summarizes the complete group count estimation procedure.
The total unsalient set
{\small \( \Omega_l^{\text{uns}} \)} is therefore partitioned into
{\small \( N_{\text{uns},l}^{\text{optimal}} \)} disjoint quantiles according to the empirical weight distribution, denoted
{\small \( \{ \Omega_l^{\text{uns}(k)} \}_{k=1}^{N_{\text{uns},l}^{\text{optimal}}} \)}.\\
Formally, {\small \( \Omega_l^{\text{uns}} = \Omega_l \setminus \Omega_l^{\text{sal}} = \bigcup_{k=1}^{N_{\text{uns},l}^{\text{optimal}}} \Omega_l^{\text{uns}(k)} \)},
with {\small \( \Omega_l^{\text{uns}(i)} \cap \Omega_l^{\text{uns}(j)} = \varnothing \)} for
{\small \( i \neq j \)}. Each subset is defined: \vspace{-0.1cm} 
{\small
\begin{equation}
\Omega_l^{\text{uns}(k)} = \{ w \in \Omega_l \mid \beta_l + z^{(k-1)} \gamma_l < |w| \leq \beta_l + z^{(k)} \gamma_l \}, \quad z^{(k)} = \Phi^{-1}\!\left(\frac{k \cdot \zeta_l^{\text{uns}} + 1}{2}\right),  \zeta_l^{\text{uns}} = 1 - \zeta_l^{\text{sal}}
\end{equation}}

\subsection{Dual-Mode Weight Quantization}
\vspace{-0.15cm}
We propose a dual-mode quantization scheme designed to minimize quantization error while incurring minimal scaling overhead. Given a weight matrix {\small \( W_l \in \mathbb{R}^{m \times n} \)}, we decompose it into a \textit{salient component}
{\small \( W_l^{\text{sal}} \in \mathbb{R}^{m \times n} \)} and multiple \textit{unsalient components}
{\small \( W_l^{\text{uns}(k)} \in \mathbb{R}^{m \times n} \)}, each corresponding to an unsalient group \( k \in \{1,...,N^{\text{optimal}}_{\text{uns}, l}\}\): \vspace{-0.2cm}
{\small
\begin{equation}
\begin{gathered}
W_l
= W_l^{\text{sal}} + \sum_{k=1}^{N_{\text{uns},l}} W_l^{\text{uns}(k)}
\\[2mm]
\begin{aligned}
\big(W_l^{\text{sal}}\big)_{ij}
&=
\begin{cases}
w_{ij}, & \text{if } w_{ij}\in \Omega_l^{\text{sal}},\\
0, & \text{otherwise},
\end{cases}
\qquad
\big(W_l^{\text{uns}(k)}\big)_{ij}
&=
\begin{cases}
w_{ij}, & \text{if } w_{ij}\in \Omega_l^{\text{uns}(k)},\\
0, & \text{otherwise}.
\end{cases}
\end{aligned}
\end{gathered}
\end{equation}
}
\begin{figure}[!t]
\centering
\vspace{-0.3cm}
\begin{minipage}[t]{0.44\linewidth}
\hrule
\vspace{1mm}
\captionsetup{type=algorithm}
\captionof{algorithm}{Graph-guided Unsalient Group Count    Estimation}
\label{alg1}
\vspace{-1mm}
\hrule
\vspace{1mm}
{\footnotesize
\begin{algorithmic}[1]
\STATE \textbf{Input:  }$\Omega_l^{\text{uns}}, N_{\text{LUT}}, S_l, K_{\text{neighbors}}$
\vspace{0.3mm}
\STATE \textbf{Output:  } $N^{\text{optimal}}_{\text{uns}, l}$

\IF{$|\Omega_l^{\text{uns}}| > S_l$}
    \STATE $\Omega_{\text{sub}} \leftarrow$ random subset of $\Omega_l^{\text{uns}}$ of size $S_l$
\ELSE
    \STATE $\Omega_{\text{sub}} \leftarrow \Omega_l^{\text{uns}}$
\ENDIF
\STATE Graph Construction
\STATE Obtain sparse $K_{\text{neighbors}}$-NN graph $\mathcal{G}$ over $\Omega_{\text{sub}}$
\STATE $\text{best\_score} \leftarrow -\infty$
\STATE $N^{\text{optimal}}_{\text{uns}, l} \leftarrow 0$

\FOR{$r \in \{ 2^{N_{\text{LUT}}-1}+1 , \ldots, 2^{N_{\text{LUT}}}-1 \}$}
    \STATE Perform spectral clustering on $\mathcal{G}$ with $r$ groups
    \STATE Compute silhouette score $s_r$ using equation~\ref{s_r}
    \IF{$s_r > \text{best\_score}$}
        \STATE $\text{best\_score} \leftarrow s_r$
        \STATE $N^{\text{optimal}}_{\text{uns}, l} \leftarrow r$
    \ENDIF
\ENDFOR

\STATE \textbf{return} $N^{\text{optimal}}_{\text{uns}, l}$
\end{algorithmic}
}
\vspace{1mm}
\hrule
\end{minipage}
\hfill
\begin{minipage}[t]{0.52\linewidth}
\hrule
\vspace{1mm}
\captionsetup{type=algorithm}
\captionof{algorithm}{Dual-Mode Quantization of Salient and Unsalient Weight Components}
\label{alg:dual_mode_quant}
\vspace{-1mm}
\hrule
\vspace{1mm}
{\footnotesize
\begin{algorithmic}[1]
\STATE \textbf{Input:} $W_l^{\text{sal}}, \{W_l^{\text{uns}(k)}\}_{k=1}^{N^{\text{optimal}}_{\text{uns},l}}, N_b, \text{iters}$
\STATE \textbf{Output:} $\hat{W}_l^{\text{sal}}, \{\hat{W}_l^{\text{uns}(k)}\}_{k=1}^{N^{\text{optimal}}_{\text{uns},l}}$

\STATE \textbf{Mode 1: Salient quantization}
\STATE $B_l^{\text{sal}} \leftarrow \text{sign}(W_l^{\text{sal}})$,\;
$\mathbf{a}^{\text{sal}} \leftarrow \mathbf{0}$

\FOR{$t=1,\ldots,\text{iters}$}
 \STATE \hspace{-0.5cm} ${a_i^{\text{sal}} \leftarrow
\dfrac{\sum_j W_{ij}^{\text{sal}} B_{ij}^{\text{sal}}}
      {\sum_j (B_{ij}^{\text{sal}})^2}, B_l^{\text{sal}} \leftarrow
\text{clip}\!\left(
\dfrac{W_l^{\text{sal}}}{\mathbf{a}^{\text{sal}}[:,\text{None}]},
-1,1\right)}$
\ENDFOR
\STATE \textbf{Discretization}
\STATE Form $2^{N_b}\!+\!1$ levels $r_{\text{levels}}$ and centers
$r_{\text{centers}}$
\STATE $B_l^{\text{sal}} \leftarrow
r_{\text{centers}}\!\big[\arg\min |B_l^{\text{sal}}-r_{\text{centers}}|\big]$
\STATE $\hat{W}_l^{\text{sal}} \leftarrow
\mathbf{a}^{\text{sal}} \odot B_l^{\text{sal}}$

\STATE \textbf{Mode 2: Unsalient binarization}
\FOR{$k=1,\ldots,N^{\text{optimal}}_{\text{uns},l}$}
    \STATE $B_l^{\text{uns}(k)}$ via~\eqref{bij_uns},
           $a_l^{\text{uns}(k)}$ via~\eqref{a_uns}
    \STATE $\hat{W}_l^{\text{uns}(k)} \leftarrow
    a_l^{\text{uns}(k)} B_l^{\text{uns}(k)}$
\ENDFOR
\STATE \textbf{return} $\hat{W}_l^{\text{sal}},
\{\hat{W}_l^{\text{uns}(k)}\}_{k}$
\end{algorithmic}
}
\vspace{1mm}
\hrule
\end{minipage}
\vspace{-0.3cm}
\end{figure}
We solve the quantization objective in Equation~\ref{opt_eq} independently for each matrix component. Similar to \cite{wang2025bi} Mode 1 applies a multi-bit quantizer (\( N_b \)-bits) to the salient weights to preserve representational precision, while Mode 2 uses a binary quantizer for each unsalient group to maximize compression.
\vspace{-0.2cm}
\paragraph{Mode 1: Multibit Quantizer for Salient Weights}
We quantize salient component \( W_l^{\text{sal}} \) into \( \hat{W}_l^{\text{sal}} \) using a single channel-wise scale per matrix.
We solve Equation~\ref{opt_eq} by assuming a diagonal scaling matrix
\( A_l^{\text{sal}} = \mathrm{diag}(a_1^{\text{sal}}, \dots, a_m^{\text{sal}}) \)
and a discretized matrix
\( B_l^{\text{sal}} \in [-1,1]^{m \times n} \).
This reduces Equation~\ref{opt_eq} to the following optimization objective:
{\small
\begin{equation} \label{sal_opt_eq}
\min_{\mathbf{a^{\text{sal}}},\, B_l^{\text{sal}}}
\left\| W_l^{\text{sal}} - \mathbf{a^{\text{sal}}} \odot B_l^{\text{sal}} \right\|_F^2,
\quad
\hat{W}_l^{\text{sal}} = \mathbf{a^{\text{sal}}} \odot B_l^{\text{sal}} .
\end{equation}}
where \( \mathbf{a^{\text{sal}}} = [a_1^{\text{sal}}, \dots, a_m^{\text{sal}}] \), \( \odot \) denotes row-wise multiplication, and
\( B_l^{\text{sal}} \) is a discretized matrix with entries drawn from a finite set of \( 2^{N_b} \) uniformly quantized levels.
We reformulate Equation~\ref{sal_opt_eq} row-wise to solve
$\min_{a_i^{\text{sal}}, b_{l,ij}^{\text{sal}}} \|w_{l,ij}^{\text{sal}} - a_i^{\text{sal}} b_{l,ij}^{\text{sal}}\|_2^2, \quad \forall j.$
The problem is non-convex since each \( b_{l,ij}^{\text{sal}} \) can only take one of \( 2^{N_b} \) discrete values in \([-1,1]\), making quantization a discrete optimization task. 
To address this, Mode 1 relaxes the constraints on \( B_l^{\text{sal}} \) and allows its entries to take continuous values in \([-1,1]\). 
Mode 1 then applies an alternating optimization scheme: first solving a quadratic program for \( \mathbf{a^{\text{sal}}} \), then updating \( B_l^{\text{sal}} \), and repeating until convergence. 
After convergence, the optimized \( B_l^{\text{sal}} \) is discretized by mapping values to the midpoints of \( 2^{N_b}+1 \) quantization levels.
\paragraph{Mode 2: Binary Quantizer for Unsalient Weights}
We quantize each unsalient component \( W_l^{\text{uns}(k)} \), \( k \in \{ 1, \dots, N_l^{\text{uns,opt}} \} \), independently to \( \hat{W}_l^{\text{uns}(k)} \) using a binary quantizer. 
Each component is mapped to a ternary matrix \( B_l^{\text{uns}(k)} \in \{-1, 0, 1\}^{m \times n} \) and a single scalar \( a_l^{\text{uns}(k)} \in \mathbb{R} \) to minimize scaling overhead. 
Rewriting Equation~\eqref{opt_eq}, we set \( A = a_l^{\text{uns}(k)} I_{m \times m} \) and \( B = B_l^{\text{uns}(k)} \), yielding the objective:
\vspace{-0.2cm}
{\small
\begin{equation} \label{uns_opt}
\begin{aligned}
\min_{a_l^{\text{uns}(k)}, B_l^{\text{uns}(k)}} 
&\left\| W_l^{\text{uns}(k)} - a_l^{\text{uns}(k)} B_l^{\text{uns}(k)} \right\|_F^2, \\
&\hat{W}_l^{\text{uns}(k)} = a_l^{\text{uns}(k)} \cdot B_l^{\text{uns}(k)}.
\end{aligned}
\end{equation}
}
\vspace{-0.1cm}
To solve for \( B_l^{\text{uns}(k)} \), Mode 2 algorithm applies an element-wise thresholding rule:
{\small
\begin{equation} \label{bij_uns} b_{l,ij}^{\text{uns}(k)*} = 
\begin{cases} \text{sign}(w_{l,ij}^{\text{uns}(k)}) & \text{if } a_l^{\text{uns}(k)} > 0, \\ -\text{sign}(w_{l,ij}^{\text{uns}(k)}) & \text{if } a_l^{\text{uns}(k)} < 0, \\ \text{0} & \text{otherwise}
\end{cases}
\end{equation} 
}
With \( B_l^{\text{uns}(k)} \) fixed, the optimal scalar is derived from the closed-form solution of Equation~\ref{uns_opt}:
\vspace{-0.1cm}
{\small
\begin{equation} \label{a_uns}
a_l^{\text{uns}(k)*} = \frac{\langle W_l^{\text{uns}(k)}, B_l^{\text{uns}(k)} \rangle}{\|B_l^{\text{uns}(k)}\|_F^2},
\end{equation}
}
We denote \( \langle W, B \rangle \) as the matrix inner product. The dual-mode quantization pipeline is summarized in Algorithm~\ref{alg:dual_mode_quant}.While we refer to unsalient quantization as \emph{binarization}, each \( B_l^{\text{uns}(k)} \) is ternary, taking values in \( \{-1, 0, 1\} \), where zeros correspond to positions outside the \( k \)-th weight group. After full quantization, the overall binary matrix \( B_l \in \{-1, 1\}^{m \times n} \) is constructed as:
{\small
\begin{equation}
    B_l = \sum_{k=1}^{N_l^{\text{uns,opt}}} B_l^{\text{uns}(k)} + \text{sign}(B_l^{\text{sal}}), \quad  B_l^{\text{uns}(k)} \in \{-1, 0, 1\}^{m \times n} 
\end{equation}
}
\subsection{Adaptive Saliency Thresholding} \label{sal_thres}
\vspace{-0.1cm}
Salient weights differ substantially from the distribution of unsalient weights. 
Misclassifying salient weights as unsalient can cause significant performance degradation~\cite{shang2023pb}, 
while selecting too many salient weights reduces compression efficiency~\cite{wang2025bi}. 
Our empirical analysis (Appendix~\ref{stat_study}) shows that the statistical properties of salient weights 
vary across model families and matrix types. 
To address this variability, we propose an adaptive method for selecting the optimal proportion of salient weights. 
We formulate saliency assignment as a numerical optimization problem that determines the optimal salient percentile 
\( \zeta_l^{\text{sal}} \) for each weight matrix \( W_l \) by minimizing the normalized reconstruction error:
{\small
\begin{equation}\label{eq:sal_opt_obj}
\begin{aligned}[t]
&\min_{\zeta_l^{\text{sal}}} \ 
\mathcal{J}(\zeta_l^{\text{sal}}; W_l, N_{\text{uns},l}^{\text{optimal}}, N_b)  = 
&\frac{\|W_l^{\text{sal}} - \mathbf{a^{\text{sal}}} \odot B_l^{\text{sal}}\|_F^2
+ \sum_{k=1}^{N_{\text{uns},l}^{\text{optimal}}}
\|W_l^{\text{uns}(k)} - a_l^{\text{uns}(k)} B_l^{\text{uns}(k)} \|_F^2}
{\|W_l\|_F^2}, \\[2pt]
&\text{s.t.}\quad
\zeta_l^{\text{sal}} \in [0, \zeta_l^{\text{sal}, \max}].
\end{aligned}
\end{equation}
}
The quantized matrices are computed using Algorithm~\ref{alg:dual_mode_quant}. 
Under the relaxed formulation of \( B_l^{\text{sal}} \) and \( B_l^{\text{uns}(k)} \), 
The objective \( \mathcal{J} \) is defined using the Frobenius norm under relaxed formulation of \( B_l^{\text{sal}} \) and \( B_l^{\text{uns}(k)} \), therefore it is convex and admits a global minimum. 
We optimize the saliency threshold using Brent’s method~\cite{brent2013algorithms} for efficient convergence. 
The resulting optimal threshold \( \zeta_l^{\text{sal,opt}} \) is then used to quantize \( W_l \).
\vspace{-0.1cm}
\subsection{Efficient Inference Runtime}
\vspace{-0.1cm}
To enable efficient assignment of scales to weight groups during inference, we propose a lightweight runtime group resolution mechanism that avoids costly block-wise saliency masks. 
Our approach pairs each weight matrix with a \textit{bit-packed} lookup table \( L_l^{\text{LUT}} \), which encodes group indices for efficient routing to quantization scales at inference time. 
For \( N_{\text{uns},l} \) unsalient clusters and one salient group, this requires only
{\small
\(
N_{\text{LUT}} = \left\lceil \log_2 (N_{\text{uns},l} + 1) \right\rceil
\)
}
bits per entry. 
The lookup table is stored in host memory and loaded layer-wise during inference. 
Due to its bit-packed representation, \( L_l^{\text{LUT}} \) incurs negligible overhead and enables simultaneous loading of metadata for multiple layers, subject to device memory constraints, thereby reducing I/O latency. 
Although \( L_l^{\text{LUT}} \) is not a traditional codebook, it effectively serves as a decoding map for dynamic weight group assignment.
\vspace{-0.2cm}
\section{Experiments}
\vspace{-0.15cm}
\subsection{Experimental Setup}
\vspace{-0.2cm}
We implement SAGE-PTQ in PyTorch \cite{paszke2019pytorch} and Huggingface \cite{wolf2019huggingface}. We run experiments on a single NVIDIA L40 46GB GPU. SAGE-PTQ operates directly on pretrained models without fine-tuning.
\vspace{-0.2cm}
\paragraph{Models and Datasets} We apply SAGE-PTQ to  LLaMA~\cite{touvron2023llama}, LLaMA-2~\cite{touvron2023llama2}, LLaMA-3~\cite{grattafiori2024llama3,dubey2024llama}, and OPT~\cite{zhang2022opt} models up to 70B parameters, and further evaluate SAGE-PTQ on recent architectures including Vicuna~\cite{chiang2023vicuna}, DeepSeek~\cite{deepseek2024}, and Qwen2.5/3~\cite{qwen2024qwen25,yang2025qwen3}. Performance is measured using perplexity, a standard metric in quantization studies, on WikiText-2~\cite{merity2016wikitext} and C4~\cite{raffel2020c4}. Perplexity is chosen for its sensitivity to distributional shifts introduced by compression. To assess generalization, we also evaluate zero-shot classification accuracy on six NLS benchmarks: PIQA~\cite{bisk2020piqa}, BoolQ~\cite{clark2019boolq}, OpenBookQA~\cite{mihaylov2018openbookqa}, Winogrande~\cite{sakaguchi2021winogrande}, ARC-c~\cite{clark2018arc}, and HellaSwag~\cite{zellers2019hellaswag}. SAGE-PTQ preserves accuracy across tasks and model families, with only minimal degradation relative to FP16 models.
\vspace{-0.2cm}
\paragraph{Baseline}
Our main baseline is BiLLM~\cite{huang2024billm}, a SoTA binarized LLM PTQ method that uses column-wise saliency selection and unsalient weight partitioning. We also compare against PB-LLM~\cite{shang2023pb}, which introduced element-wise saliency-based binarization. In addition, we include multibit PTQ baselines, including 2-/3-bit GPTQ~\cite{frantar2023gptq} and 3-bit AWQ \cite{lin2024awq} methods.
\vspace{-0.15cm}
\subsection{Results}
\vspace{-0.1cm}
\paragraph{Evaluation of Language Generation Tasks}
We evaluate SAGE-PTQ across multiple LLMs and sizes, comparing against binarized~\cite{huang2024billm,shang2023pb} and multibit PTQ methods~\cite{lin2024awq,frantar2023gptq}.  Language generation quality is measured by perplexity. We explicitly account for scaling overhead during inference, an aspect often neglected in prior work. 

Table~\ref{tab1} presents SAGE-PTQ results on OPT models from 1.3B to 66B, using 4-bit precision for salient weights in outlier-heavy first layer (Appendix \ref{stat_study}) and 2-bit salient precision in later layers. Experiments are conducted with lookup constraints $N_{\text{LUT}}=3$ and $N_{\text{LUT}}=4$, comparing WikiText-2 perplexity against BiLLM, PB-LLM,GPTQ and AWQ. The 4-bit lookup setting matches baseline storage, while the 3-bit setting highlights robustness under tighter storage constraints. Recall that lookup values are decoded once per layer and do not add to device memory overhead. SAGE-PTQ achieves the best perplexity across all model sizes and lookup settings. We enforce a maximum of 10\% salient weights, our adaptive saliency thresholding \ref{sal_thres} yields an average of 7\%, resulting in a binarized weight precision of 1.07. Crucially, SAGE-PTQ requires only 0.009 bits per weight for scale storage, far below all competing methods.
\begin{table*}[t]
\captionsetup{justification=raggedright}
\caption{Perplexity results on WikiText2 for the OPT family.
We compare FP16 baselines, 3-bit PTQ methods (GPTQ, AWQ),
sub-2-bit methods (GPTQ-2bit, PB-LLM (10\%), BiLLM), and SAGE-PTQ.
SAGE-PTQ consistently achieves competitive or lower perplexity.}
\label{tab1}
\begin{center}
\begin{small}
\footnotesize
\setlength{\tabcolsep}{2.5mm}
\renewcommand{\arraystretch}{0.9}
\vspace{1ex}
\vspace{-0.3cm}
\begin{tabular}{lcccccccc}
\toprule
\textbf{Method} & \textbf{Weight bits} & \textbf{Scale bits} &
\textbf{1.3B} & \textbf{2.7B} & \textbf{6.7B} & \textbf{13B} & \textbf{30B} & \textbf{66B} \\
\midrule
\textbf{FP16}          & 16.00 & 16.00 & 14.62 & 12.47 & 10.86 & 10.13 & 9.56 & 9.34 \\
\addlinespace[0.6ex]
\hdashline
\addlinespace[0.6ex]
GPTQ           & 3.00  & 0.25  & 20.97 & 16.88 & 14.86 & 11.61 & 10.27 & 14.16 \\
AWQ           & 3.00  & 0.25  & 16.32 & 13.58 & 11.39 & 10.56 & 9.77  & 9.62  \\
\addlinespace[0.6ex]
\hdashline
\addlinespace[0.6ex]
GPTQ           & 2.00 & 0.25  & 115.17 & 61.59  & 50.19 & 21.36 & 15.71 & 82.10 \\
PB-LLM (10\%)          & 1.70 & 0.50  & 265.52 & 124.35 & 105.16 & 81.92 & 25.14 & 29.09 \\
BiLLM                  & 1.11 & 1.00  & 69.97  & 49.55  & 35.36 & 18.82 & 12.71 & 12.06 \\
\addlinespace[0.7ex]
\hdashline
\addlinespace[0.7ex]
\textbf{SAGE-PTQ (N\textsubscript{LUT}=3)} & \textbf{1.07} & \textbf{0.009} &
\textbf{17.90} & \textbf{14.60} & \textbf{13.87} & \textbf{11.33} & \textbf{10.09} & \textbf{11.35} \\
\textbf{SAGE-PTQ (N\textsubscript{LUT}=4)} & \textbf{1.07} & \textbf{0.009} &
\textbf{15.63} & \textbf{13.23} & \textbf{11.59} & \textbf{10.91} & \textbf{9.82}  & \textbf{11.31} \\
\bottomrule
\end{tabular}
\end{small}
\end{center}
\vskip -0.1in
\vspace{0.15cm}
\end{table*}
\begin{table*}[t]
\captionsetup{justification=raggedright}
\caption{WikiText2 perplexity results of GPTQ, PB-LLM (10\%), BiLLM and SAGE-PTQ on LLaMA-1-2-3,
SAGE-PTQ consistently achieves lower perplexity.
*: LLaMA has a (7B,65B) variants; LLaMA-2 has (7B,70B) variants and LLaMA-3 has a (8B,70B) variants.}
\label{tab2}
\begin{center}
\begin{small}
\footnotesize
\setlength{\tabcolsep}{2.9mm}
\renewcommand{\arraystretch}{0.8}
\vspace{1ex}
\vspace{-0.35cm}
\begin{tabular}{l l c c c c c c}
\toprule
\textbf{Model} & \textbf{Method} & \textbf{Weight bits} & \textbf{Scale bits} &
\textbf{7B/8B*} & \textbf{13B} & \textbf{30B} & \textbf{65B / 70B*} \\
\midrule
\multirow{4}{*}{\textbf{LLaMA-1}}
  & GPTQ             & 2.00 & 0.25  & 152.31 & 20.44  & 13.01 & 8.78  \\
  & PB-LLM (10\%)    & 1.70 & 0.50  & 102.36 & 36.60  & 33.67 & 12.53 \\
  & BiLLM            & 1.09 & 1.00  & 35.04  & 15.14  & 10.52 & 8.49  \\
\addlinespace[0.5ex]
\cdashline{2-8}
\addlinespace[0.5ex]
  & \textbf{SAGE-PTQ} & \textbf{1.03} & \textbf{0.004} &
    \textbf{5.99} & \textbf{5.35} & \textbf{5.97} & \textbf{4.93} \\
\midrule
\multirow{4}{*}{\textbf{LLaMA-2}}
  & GPTQ             & 2.00 & 0.25  & 60.45  & 19.70  & NA    & 9.12  \\
  & PB-LLM (10\%)    & 1.00 & 0.50  & 69.20  & 151.09 & NA    & 28.37 \\
  & BiLLM            & 1.08 & 1.00  & 32.48  & 16.77  & NA    & 8.41  \\
\addlinespace[0.5ex]
\cdashline{2-8}
\addlinespace[0.5ex]
  & \textbf{SAGE-PTQ} & \textbf{1.03} & \textbf{0.004} &
    \textbf{5.87} & \textbf{14.49} & \textbf{NA} & \textbf{7.51} \\
\midrule
\multirow{4}{*}{\textbf{LLaMA-3}}
  & GPTQ             & 2.00 & 0.25  & 480.43 & NA     & NA    & 82.23 \\
  & PB-LLM (10\%)    & 1.70 & 0.50  & 73.08  & NA     & NA    & 22.96 \\
  & BiLLM            & 1.09 & 1.00  & 55.80  & NA     & NA    & 66.30 \\
\addlinespace[0.5ex]
\cdashline{2-8}
\addlinespace[0.5ex]
  & \textbf{SAGE-PTQ} & \textbf{1.03} & \textbf{0.004} &
    \textbf{6.74} & \textbf{NA} & \textbf{NA} & \textbf{9.16} \\
\bottomrule
\end{tabular}
\end{small}
\end{center}
\vskip -0.1in
\vspace{-0.3cm}
\end{table*}

Table~\ref{tab2} shows SAGE-PTQ evaluation on LLaMA-1, LLaMa-2 and LLaMA-3 up to 70B, under lookup constraint $N_{\text{LUT}}=4$. For models under 30B, only 1\% of weights are marked salient and quantized with 4-bit precision in all layers due to outlier sensitivity. SAGE-PTQ achieves a 1.03 average bitwidth and 0.004 bits of scaling overhead, outperforming all baselines in perplexity. 
\vspace{-0.05cm}

The scaling overhead is computed by accounting for the number of FP16 scale values required to decode weights at inference. GPTQ and PB-LLM use block-wise min-max scaling on 128-column blocks. GPTQ and AWQ stores one min and one max column (2 FP16 columns), contributing 0.25 bits per weight. PB-LLM stores 4 columns (2 for salient, 2 for unsalient), totaling 0.5 bits. BiLLM further adds residual scaling and uses 8 columns per block (4 for salient, 4 for unsalient), requiring 1.0 bit. In contrast, SAGE-PTQ uses a single FP16 scale for the entire matrix and one scalar per unsalient cluster, avoiding position-based heuristics and yielding negligible scale cost.

\vspace{-0.05cm}
We evaluate SAGE-PTQ with \(N_{\text{LUT}}=4\) on recent similarly sized architectures, including Vicuna-7B, DeepSeek-7B-Base, LLaMA-3-8B, Qwen2.5-7B, and Qwen3-8B-Base, using WikiText2 and C4. Table~\ref{tab3_diverse_ppl} shows that SAGE-PTQ consistently outperforms BiLLM on both datasets, reducing perplexity by 77.6\% and 71.7\% on average, respectively. Models with long-tailed weight distributions, such as LLaMA-3 and Vicuna, benefit from 4-bit salient quantization at a 1\% saliency threshold, while Qwen and DeepSeek perform well with 2-bit salient quantization at a 10\% threshold. We evaluate larger models and additional architectures on other datasets~\cite{marcus1994ptb}; provided in Appendix~\ref{experi}.
\vspace{-0.3cm}
\begin{table}[!h]
\centering
\footnotesize
\setlength{\tabcolsep}{7.2pt}
\caption{Perplexity comparison between BiLLM and \textsc{SAGE-PTQ} on WikiText2 and C4. \textsc{SAGE-PTQ} achieves lower perplexity than BiLLM on both datasets.}
\label{tab3_diverse_ppl}
\vspace{0.15cm}
\begin{tabular}{llccccc}
\toprule
\textbf{Dataset} & \textbf{Method }
& \textbf{Vicuna-7B} 
& \textbf{DeepSeek-7B }
& \textbf{LLaMA-3-8B} 
& \textbf{Qwen-2.5-7B} 
& \textbf{Qwen-3-8B} \\
\midrule
\multirow{2}{*}{\textbf{WikiText2}}
& BiLLM 
& 33.00 & 19.83 & 55.80 & 41.74 & 48.20 \\
& \textbf{SAGE-PTQ} 
& \textbf{7.27} & \textbf{7.18} & \textbf{6.75} & \textbf{8.45} & \textbf{8.40} \\
\midrule
\multirow{2}{*}{\textbf{C4}}
& BiLLM 
& 28.75 & 33.33 & 64.89 & 53.07 & 41.40 \\
& \textbf{SAGE-PTQ} 
& \textbf{9.90} & \textbf{11.22} & \textbf{10.30} & \textbf{14.34} & \textbf{14.50} \\
\bottomrule
\end{tabular}
\end{table}
\begin{table}[t]
\centering
\footnotesize
\setlength{\tabcolsep}{8.4pt}
\caption{Zero-shot accuracy comparison of FP16, BiLLM, and \textsc{SAGE-PTQ} across language understanding benchmarks. \textsc{SAGE-PTQ} achieves higher accuracy than BiLLM across all benchmarks}
\label{tab:zeroshot_results}
\vspace{1ex}
\begin{tabular}{llcccccc}
\toprule
\textbf{Model} & \textbf{Method} 
& \textbf{PIQA} 
& \textbf{BoolQ} 
& \textbf{OBQA} 
& \textbf{WinoGrande} 
& \textbf{ARC-e} 
& \textbf{HellaSwag} \\
\midrule
\multirow{3}{*}{\textbf{OPT-6.7B}}
& FP16 
& 75.68 & 68.66 & 31.60 & 60.38 & 31.10 & 64.09 \\
& BiLLM 
& 58.60 & 62.20 & 29.00 & 51.50 & 23.90 & 31.90 \\
\cdashline{2-8}
& \textbf{SAGE-PTQ} 
& \textbf{73.94} & \textbf{64.71} & \textbf{29.20} & \textbf{58.56} & \textbf{26.76} & \textbf{56.68} \\
\midrule
\multirow{3}{*}{\textbf{LLaMA-3-8B}}
& FP16 
& 80.25 & 83.36 & 40.60 & 71.19 & 43.81 & 77.13 \\
& BiLLM 
& 57.51 & 65.93 & 30.14 & 58.12 & 30.75 & 31.75 \\
\cdashline{2-8}
& \textbf{SAGE-PTQ} 
& \textbf{78.78} & \textbf{74.43} & \textbf{38.20} & \textbf{66.54} & \textbf{39.80} & \textbf{73.15} \\
\midrule
\multirow{3}{*}{\textbf{Qwen-3-8B}}
& FP16 
& 78.94 & 83.03 & 42.40 & 67.48 & 47.83 & 74.38 \\
& BiLLM 
& 62.50 & 63.90 & 37.10 & 53.00 & 43.80 & 34.30 \\
\cdashline{2-8}
& \textbf{SAGE-PTQ} 
& \textbf{77.31} & \textbf{80.28} & \textbf{41.00} & \textbf{64.33} & \textbf{46.49} & \textbf{71.61} \\
\bottomrule
\end{tabular}
\vspace{-0.35cm}
\end{table}
\vspace{-0.25cm}
\paragraph{Zero-shot Results}
We evaluate SAGE-PTQ under 4-bit lookup constraints on six zero-shot NLU QA benchmarks: PIQA, BoolQ, OBQA, Winogrande, ARC-e, and HellaSwag, using OPT-6.7B, LLaMa-3-8B and Qwen3-8B-base models. Table \ref{tab:zeroshot_results} shows that SAGE-PTQ matches FP16 accuracy on most tasks, outperforming BiLLM by an average of 5\% across all benchmarks.
\vspace{-0.35cm}
\paragraph{Memory Footprint} A primary goal of LLM PTQ is to reduce GPU memory required for inference. We evaluate SAGE-PTQ memory footprint, avoiding untracked device overhead by offloading lookup metadata to host memory and storing only weights and scales on GPU. 

We compare against FP16 for active inference memory and FP32 for idle disk storage.
We assume best-case lookup overhead for BiLLM, as BiLLM group retrieval mechanism is unspecified. BiLLM uses block-wise partitioning: with 32 blocks of 128 columns for 7B models and up to 64 blocks of 128 columns for 70B models.

 \begin{wrapfigure}{r}{0.58\textwidth}
\vspace{-1.0em}
\centering
\includegraphics[
    width=\linewidth,
    height=0.153\textheight,
    keepaspectratio
]{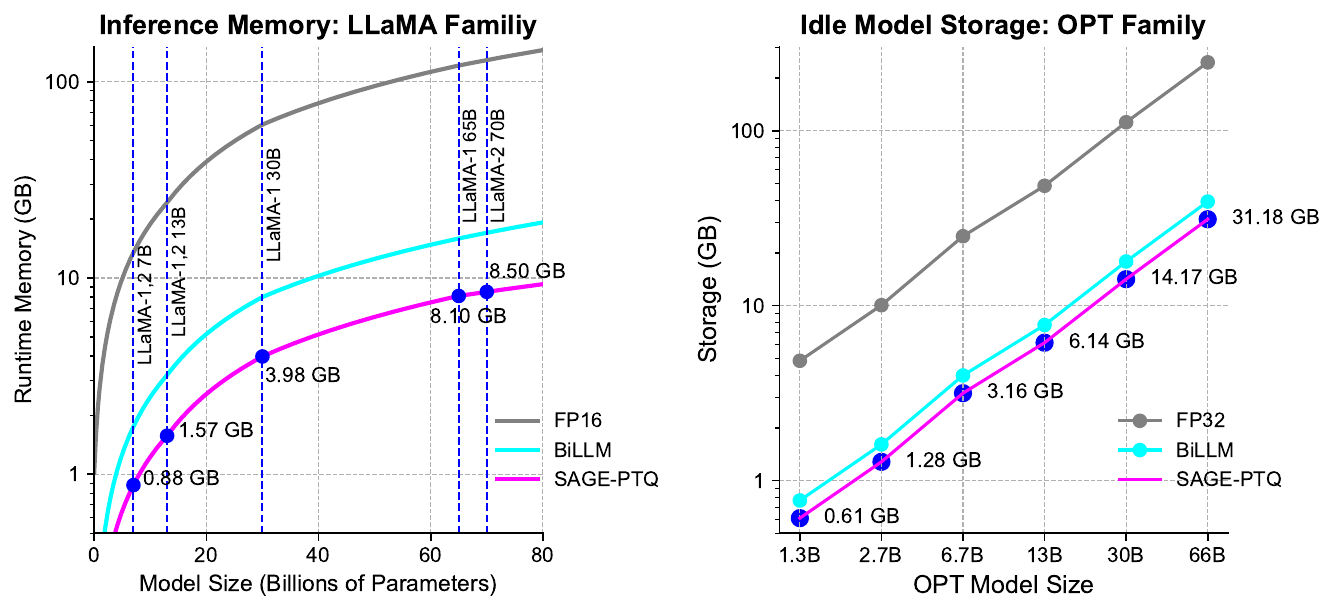}
\caption{Memory footprint of SAGE-PTQ versus BiLLM and ful-precision for LLaMA and OPT models. SAGE-PTQ outperforms BiLLM by using 50\% less GPU memory during inference while also reduces idle mode storage.}
\label{fig:storage}
\vspace{-2.0em}
\end{wrapfigure}
For fixed block positions and three weight groups per block, the best-case BiLLM lookup requires ${ \lceil \log_2(3) \rceil = 2 }$ average bits per weight matrix, included alongside weight and scale bits.
We measure runtime GPU memory for LLaMA-1 and LLaMA-2, and idle host memory for OPT. Figure~\ref{fig:storage} shows that SAGE-PTQ achieves SoTA efficiency by 50.3\% reduction in GPU memory, averaging 1.038 bits for weights  and scales, while BiLLM requires an average of 2.08 bits. For OPT models; SAGE-PTQ still lowers idle storage by 20.7\%.

\vspace{-0.3cm}

\paragraph{End-to-end Inference Latency}  
We report the inference latency of SAGE-PTQ through a case study on LLaMA-2-70B (140\,GB in FP16) deployed on a single NVIDIA L40 GPU (46\,GB). Since the model has 80 layers and cannot fully reside in GPU memory, standard FP16 deployment loads a small band of layers (e.g., $B_{FP16}=5$) from CPU RAM per token, introducing significant I/O latency.
With SAGE-PTQ and a 3-bit lookup budget, the quantized model fits in $\sim$12.7\,GB, with $\sim$3\,GB for KV-cache during prefill for a context length of 2048. During decoding, SAGE-PTQ enables on-the-fly dequantization of a larger band ($B_Q=16$) using lookup metadata, avoiding full layer transfers.

We measure layer preprocessing time and per-token generation latency in Table~\ref{tab3}. SAGE-PTQ introduces moderate dequantization overhead but substantially reduces I/O latency by avoiding full layer transfers. With in-GPU preprocessing and parallel band dequantization, SAGE-PTQ achieves a 1.5$\times$ speedup in per-token decoding time for LLaMA-2-70B, making it suitable for memory-constrained scenarios. For a context length \(L_0\), band I/O overhead \(d_Q\), and dequantization cost \(D\), the per-token latency captures the trade-off between memory efficiency and latency. For short generations of length \(T \ll L_0\), the total quantized latency is
${\approx O\left(T \cdot \left(N_L/B_Q\right)(d_Q + D)\right)}$,
showing approximately linear scaling with sequence length \(T\) and hardware overheads.

\paragraph{Quantization Time} 
We evaluate SAGE-PTQ quantization efficiency by measuring the total time required to quantize full models on a single NVIDIA L40 GPU. Table~\ref{tab5} shows that SAGE-PTQ achieves faster quantization times compared to BiLLM, despite involving multiple stages. SAGE-PTQ is device-optimized and quantizes a 66B parameter model ($\sim$ 140GB) in just 1.5 hours.
\begin{table}[!t]
\vspace{1ex}
\centering
\footnotesize
\setlength{\tabcolsep}{3.5pt}
\renewcommand{\arraystretch}{0.85}

\begin{minipage}[t]{0.48\linewidth}
\centering
\caption{Inference latency breakdown for SAGE-PTQ and unquantized LLaMA-2-70B.}
\label{tab3}
\vspace{0.3ex}
\begin{tabular}{@{}lccc@{}}
\toprule
\textbf{Method} 
& \textbf{Layer I/O} 
& \textbf{Band dequant.} 
& \textbf{Token latency} \\
\midrule
FP16 
& 47.07\,ms 
& -- 
& 972.1\,ms \\
\textbf{SAGE-PTQ} 
& 10.46\,ms 
& 75.52\,ms 
& \textbf{646.3\,ms} \\
\bottomrule
\end{tabular}
\end{minipage}
\hfill
\begin{minipage}[t]{0.48\linewidth}
\centering
\caption{Full SAGE-PTQ runtime for quantizing the OPT family.}
\label{tab5}
\vspace{0.3ex}
\begin{tabular}{@{}lcccccc@{}}
\toprule
\textbf{Method} 
& \textbf{1.3B} 
& \textbf{2.7B} 
& \textbf{6.7B} 
& \textbf{13B} 
& \textbf{30B} 
& \textbf{66B} \\
\midrule
BiLLM 
& 5.1m & 8.9m & 15.8m & 26.4m & 51.9m & 1.7h \\
\textbf{SAGE-PTQ} 
& \textbf{4.2m} 
& \textbf{6.6m} 
& \textbf{10.2m} 
& \textbf{19.5m} 
& \textbf{43.9m} 
& \textbf{1.5h} \\
\bottomrule
\end{tabular}
\end{minipage}

\vspace{1ex}
\vspace{-0.3cm}
\end{table}
\vspace{-0.2cm}
\subsection{Ablation Study}
\begin{figure}[h]
    \centering
    \begin{minipage}[t]{0.42\linewidth}
        \centering
        \includegraphics[
            width=\linewidth,
        ]{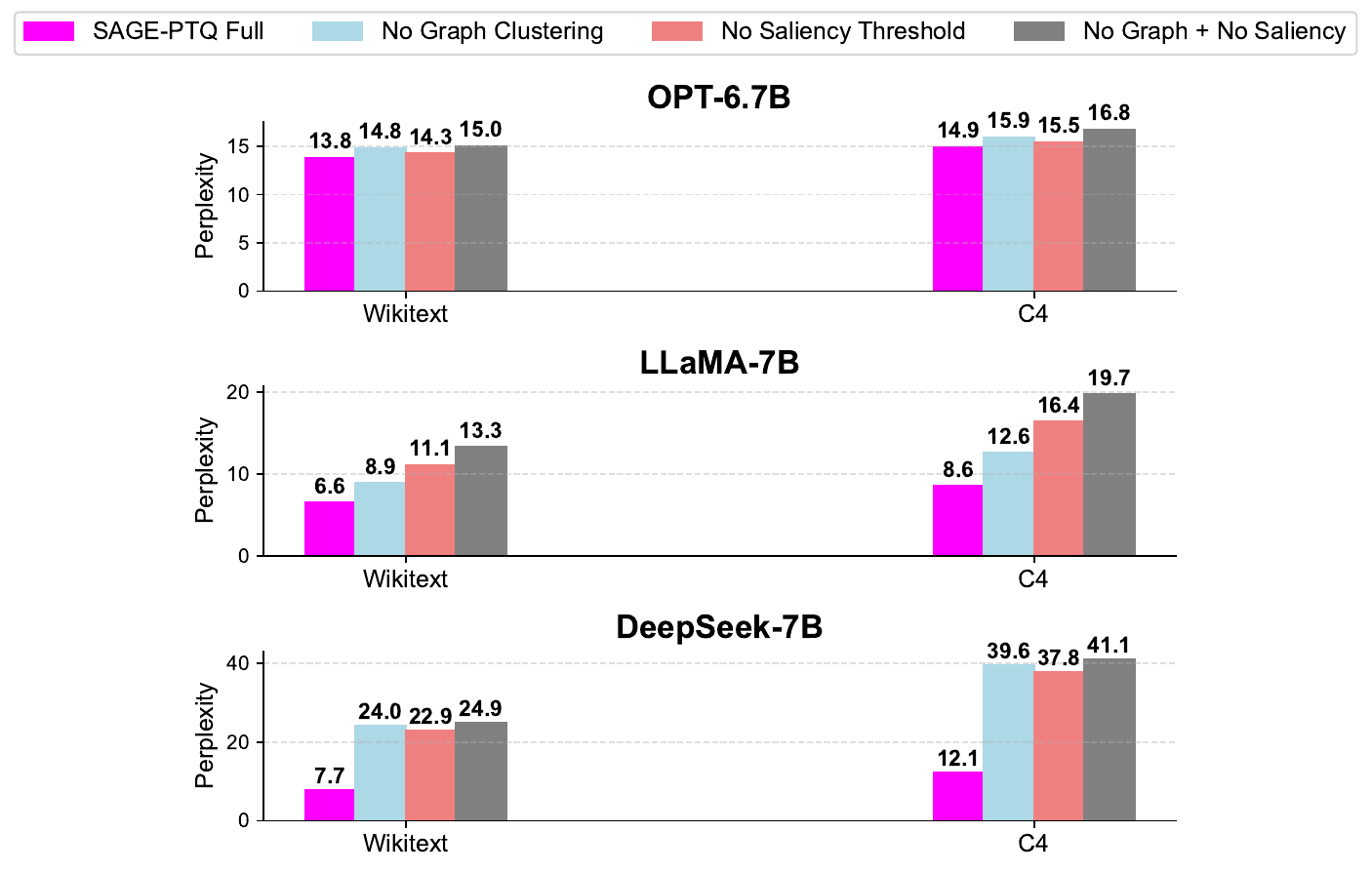}
        \caption{Ablation results of SAGE-PTQ on OPT 6.7B, LLaMA 7B, and DeepSeek 7B. Full pipeline achieves best perplexity.}
        \label{fig:ablation}
    \end{minipage}
    \hfill
    \begin{minipage}[t]{0.49\linewidth}
        \centering
        \includegraphics[
            width=\linewidth
        ]{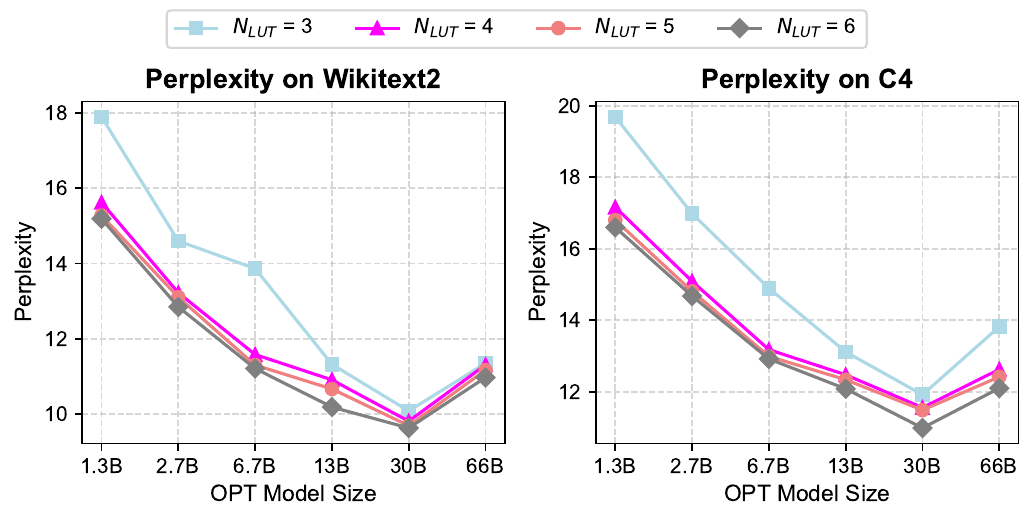}
        \caption{OPT models perplexity on WikiText2 and C4 under different lookup sizes. A 4-bit lookup balances performance-storage trade-off.}
        \label{fig:ablation2}
    \end{minipage}
\end{figure}

\vspace{-0.2cm}
\textsc{SAGE-PTQ} combines adaptive saliency selection and graph-guided weight partitioning, we evaluate the effectiveness of method pipeline components. We also address balancing memory footprint and SAGE-PTQ performance. Appendix \ref{sal_ablation} presents more ablation analysis on saliency metric selection.
\vspace{-0.5cm}
\paragraph{Effectiveness of Saliency Thresholding and Graph-guided Modeling} We ablate adaptive saliency thresholding and graph-guided group estimation in SAGE-PTQ while keeping the dual quantization fixed. We evaluate on WikiText2 and C4 using LLaMA-7B, OPT-6.7B, and DeepSeek-7B to cover diverse architectures and weight distributions. Figure~\ref{fig:ablation} shows that removing either stage increases perplexity. The full method achieves the lowest perplexity, with ~2.4$\times$ improvement on outlier-heavy models (LLaMA,DeepSeek), higher than on OPT-6.7B, which has a more regular weight distribution.
\vspace{-0.52cm}
\paragraph{Tradeoff between Lookup Overhead and Quantization Performance}
\textsc{SAGE-PTQ} uses a novel lookup strategy to recover weight-group information at inference. We study the effect of lookup size on performance. As shown in Figure~\ref{fig:ablation2}, increasing the lookup size improves perplexity on WikiText2 and C4 for the OPT family by enabling finer group granularity and more accurate scale assignment. Results for other model families are provided in Appendix~\ref{lut_ablation}. Notably, a 4-bit lookup closely matches higher-bit settings, capturing salient weight structure with low overhead.
\vspace{-0.5cm}
\section{Conclusion, Limitations and Future Work}
\vspace{-0.1cm}
We introduced \textsc{SAGE-PTQ}, a PTQ framework for LLMs that addresses key limitations of prior ultra-low-bit methods by jointly considering hidden scaling overhead, weight partitioning, and quantization error. Through graph-guided group-count estimation and adaptive saliency thresholding, \textsc{SAGE-PTQ} improves perplexity and inference efficiency across diverse model families and scales.
Extensive experiments show that \textsc{SAGE-PTQ} consistently outperforms strong baselines such as BiLLM and PB-LLM, achieving up to $5\times$ lower perplexity with only $1.03$ average bits per weight and $0.004$ scaling bits. Compared to BiLLM, it reduces runtime GPU memory by over 50\% on LLaMA models while preserving near-FP16 zero-shot accuracy.
A limitation of \textsc{SAGE-PTQ} is that it performs weight-only quantization and does not address activation quantization. Future work will extend the framework to weight-activation quantization and broaden evaluation to recent architectures, including MoE-based LLMs. Overall, \textsc{SAGE-PTQ} is a promising approach for deploying LLMs on resource-constrained platforms. The code will be released soon.

{\small
\bibliography{references}
\bibliographystyle{unsrt}
}
\newpage
\appendix

\section{Statistical Analysis of Weight Matrices and Outlier Pattern Detection}
\label{stat_study}
Efficient post-training quantization for large language models (LLMs) demands careful modeling of weight matrix statistics, including \textit{saliency}, \textit{sparsity}, and \textit{outliers}. \textsc{SAGE-PTQ} is a saliency-aware framework designed from empirical analysis across model families and layers. It enables \textit{mixed-precision quantization} by preserving higher precision for salient weights while reducing overall bit usage. This data-driven design ensures adaptability to diverse architectures and yields improved trade-offs between compression and accuracy across LLMs.

We begin by analyzing the OPT model family, focusing on OPT-6.7B as a representative intermediate-scale model with 32 transformer layers. For each layer, we study the value distribution of self-attention projection weight matrices---\textit{query}, \textit{key}, \textit{value}, and \textit{output}---each of size $4096 \times 4096$. We track how these distributions evolve from early (layer~0) to mid and final layers (e.g., layer~30). Similar analyses are conducted on LLaMA-1 7B and DeepSeek-Base 7B to compare distributional patterns across model families. For each weight matrix, we compute histograms of individual values and fit a Gaussian curve using the empirical mean and standard deviation. We calculate the Kullback--Leibler (KL) divergence between the histogram and fitted Gaussian to quantify deviation from normality. This tests whether the law of large numbers (LLN) applies and reveals layers with statistical anomalies or salient outliers. These observations guide the design of our quantization method, ensuring adaptability to model-specific and layer-wise weight characteristics.

\begin{figure*}[!t]
    \centering
    \includegraphics[width=0.95\textwidth, height=7.8cm]{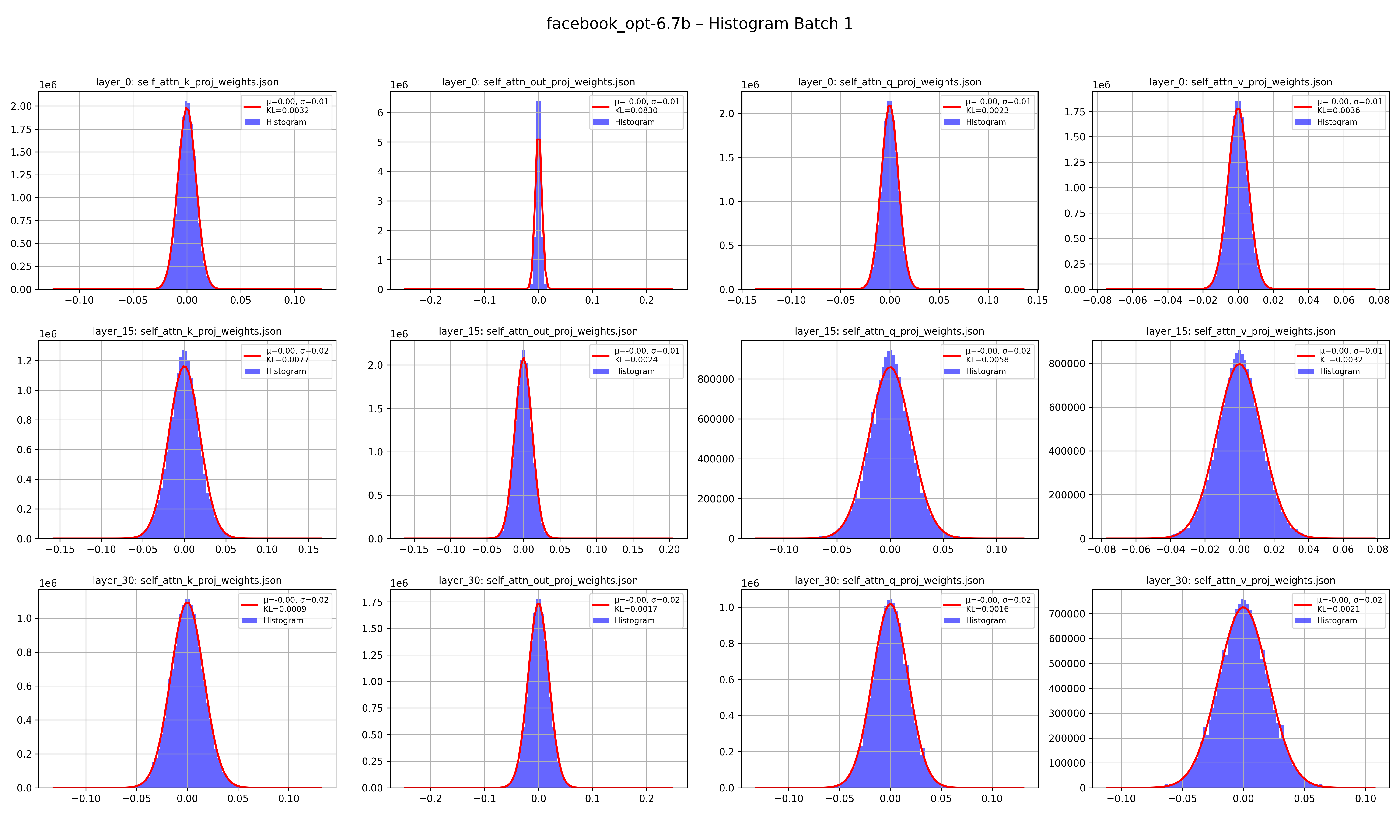}
    \caption{Distribution of self-attention projection layer weights in OPT-6.7B across selected layers (0,15,30). All layers follow a near-Gaussian trend with low KL divergence, except for the \texttt{out\_proj} matrix in the first layer, which shows significant deviation.}
    \label{opt_stat}
\end{figure*}

\paragraph{OPT-6.7B:} As shown in figure \ref{opt_stat}, Most layers in the OPT-6.7B model exhibit weight distributions that closely match a Gaussian curve, with KL divergence typically below 0.1. The long tail of the histogram contains salient weights, and the tail length varies across layers. The first layer deviates significantly from Gaussianity, suggesting early-layer adaptation. Standard deviation of weights varies across layers, indicating nonuniformity in weight distribution statistics.

\begin{figure*}[!t]
    \centering
    \includegraphics[width=0.95\textwidth, height=7.8cm]{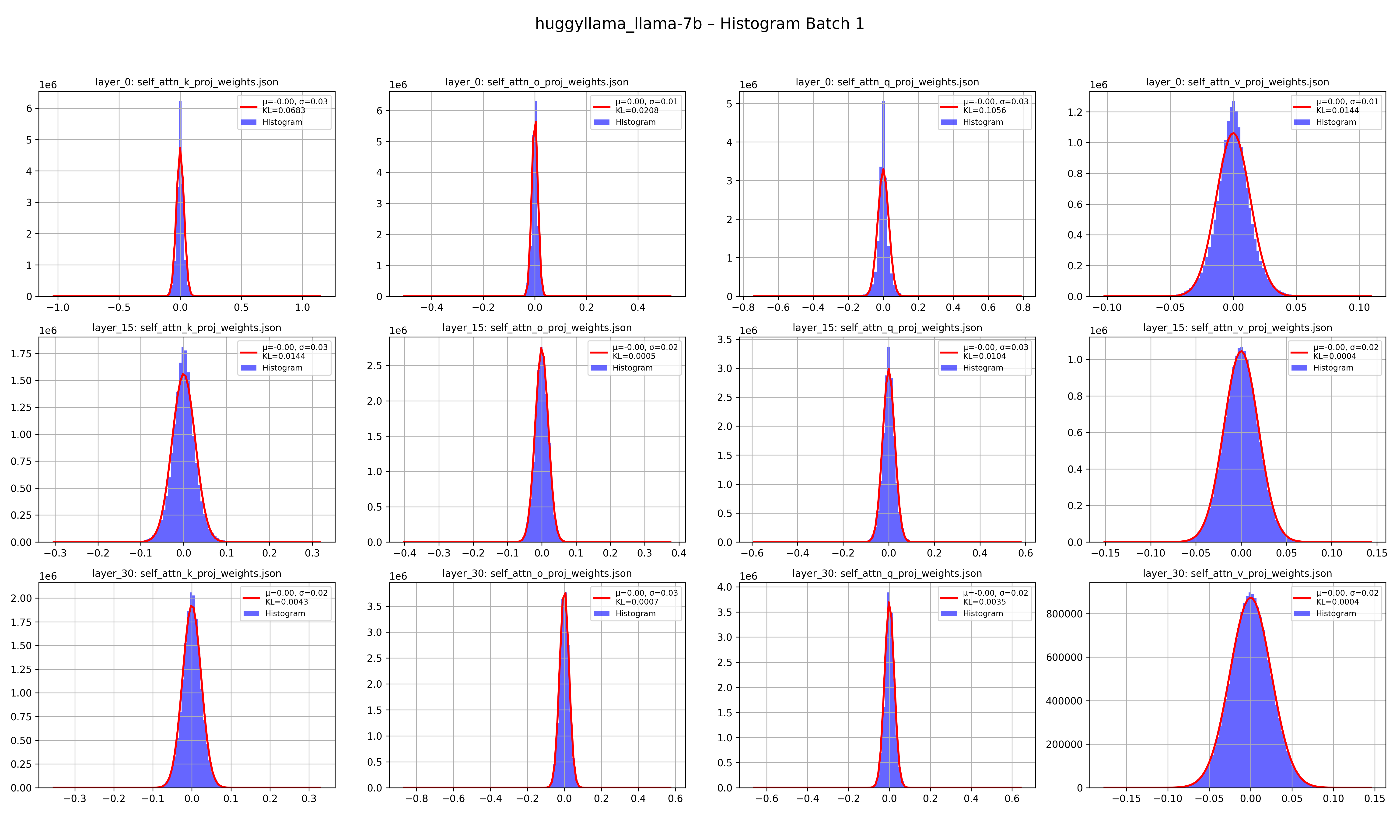}
    \caption{Distribution of self-attention projection layer weights in LLaMa-1 7B  across Self-Attention selected layers (0,15,30). Observed Gaussian Trends with Sparse High-Magnitude Outliers.}
    \label{llama_stat}
\end{figure*}

\paragraph{LLaMA-1 7B:} In figure \ref{llama_stat}, The LLaMA-1 7B model displays a small number of high-magnitude outliers. While the bulk of the weight distribution remains approximately Gaussian, sharp deviations are again observed in the first layer. The sensitivity of outliers to quantization and the variation in standard deviation across layers suggest the need for adaptive, outlier-aware weight partitioning.

\begin{figure*}[!t]
    \centering
    \includegraphics[width=0.95\textwidth, height=7.8cm]{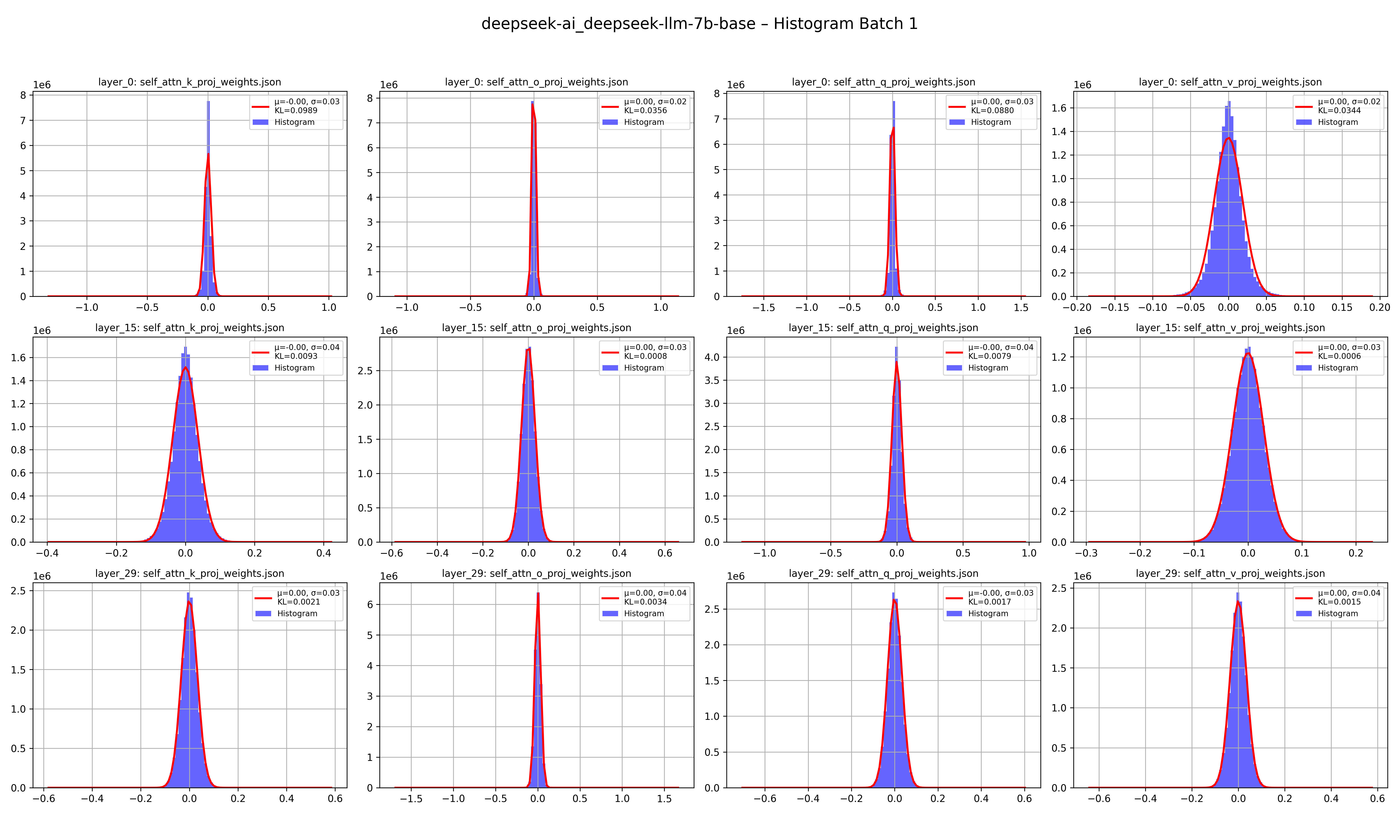}
    \caption{Self-Attention Weight Distributions in DeepSeek-7B of selected layers (0,15,29): Non-Gaussian Patterns with Widespread High-Magnitude Outliers.}
    \label{deepseek_stat}
\end{figure*}

\paragraph{DeepSeek-7B:} Figure \ref{deepseek_stat} shows that the DeepSeek-7B model consistently deviates from Gaussianity across all layers. It contains high-magnitude outliers in long-tailed distributions. Inlier weights exhibit relatively small but layer-dependent standard deviations. These observations suggest that fewer weight groups are required, with more bits allocated to salient weights.

\begin{itemize}
    \item OPT-6.7B layers are mostly Gaussian, except for strong deviation in the first layer.
    \item LLaMA-1 7B contains large-magnitude outliers and early-layer deviations.
    \item DeepSeek-7B shows persistent non-Gaussian behavior and long-tailed outliers.
    \item Standard deviation of inlier weights varies significantly across layers.
    \item Inlier weight distributions are layer-specific and inconsistent across models.
\end{itemize}
\textbf{Quantization Strategy Requirements:}
\begin{itemize}
    \item Detect and isolate outliers before quantization.
    \item Avoid fixed-position or global threshold-based partitioning.
    \item Assign higher precision to salient weights to prevent quantization distortion.
    \item Use affinity-based partitioning for unsalient weights instead of value range alone.
    \item Apply a graph-based partitioning method to model intra-layer similarity.
    \item Adapt group count and saliency thresholds dynamically per layer.
\end{itemize}

These insights motivated the design of \textsc{SAGE-PTQ}, including:
\begin{itemize}
    \item Saliency filtering to isolate high-impact weights,
    \item Graph-based modeling to group inliers by affinity,
    \item Adaptive saliency thresholding to prevent outlier contamination within unsalient groups.
\end{itemize}
\section{Extended Experimental Results} 
\label{experi}
\subsection{Language Generation Tasks}
To rigorously evaluate the generalization and robustness of \textsc{SAGE-PTQ} across diverse model families and scales, we extend our analysis to larger models beyond the 7B parameter range. We report perplexity on three benchmark datasets: WikiText2, PTB, and C4, and compare our results against BiLLM. We only compare against BiLLM since it achieves strong performance at comparable bitwidths, particularly in binarized settings.
All models are evaluated under a lookup table constraint of $N_{\text{LUT}} = 4$ bits. For OPT models, which exhibit regular Gaussian weight distributions with consistent outlier behavior (as detailed in Appendix A), we assign $N_b = 2$ bits to salient weights and $N_b = 4$ for the first layer to account for its deviation from normality. A saliency threshold of 10\% is used. For LLaMA and Vicuna models, where outliers are rare but extreme, we apply a stricter 1\% saliency threshold. \textsc{SAGE-PTQ} achieves an average bitwidth of 1.07 for OPT and 1.03 for LLaMA/Vicuna, outperforming BiLLM on all datasets. Our method delivers new state-of-the-art results with perplexity scaling factors as low as 0.004 for LLaMA/Vicuna and 0.009 for OPT.
\begin{figure*}[!b]
    \label{extended}
    \centering
    \includegraphics[width=0.9\textwidth,height=7.5cm]{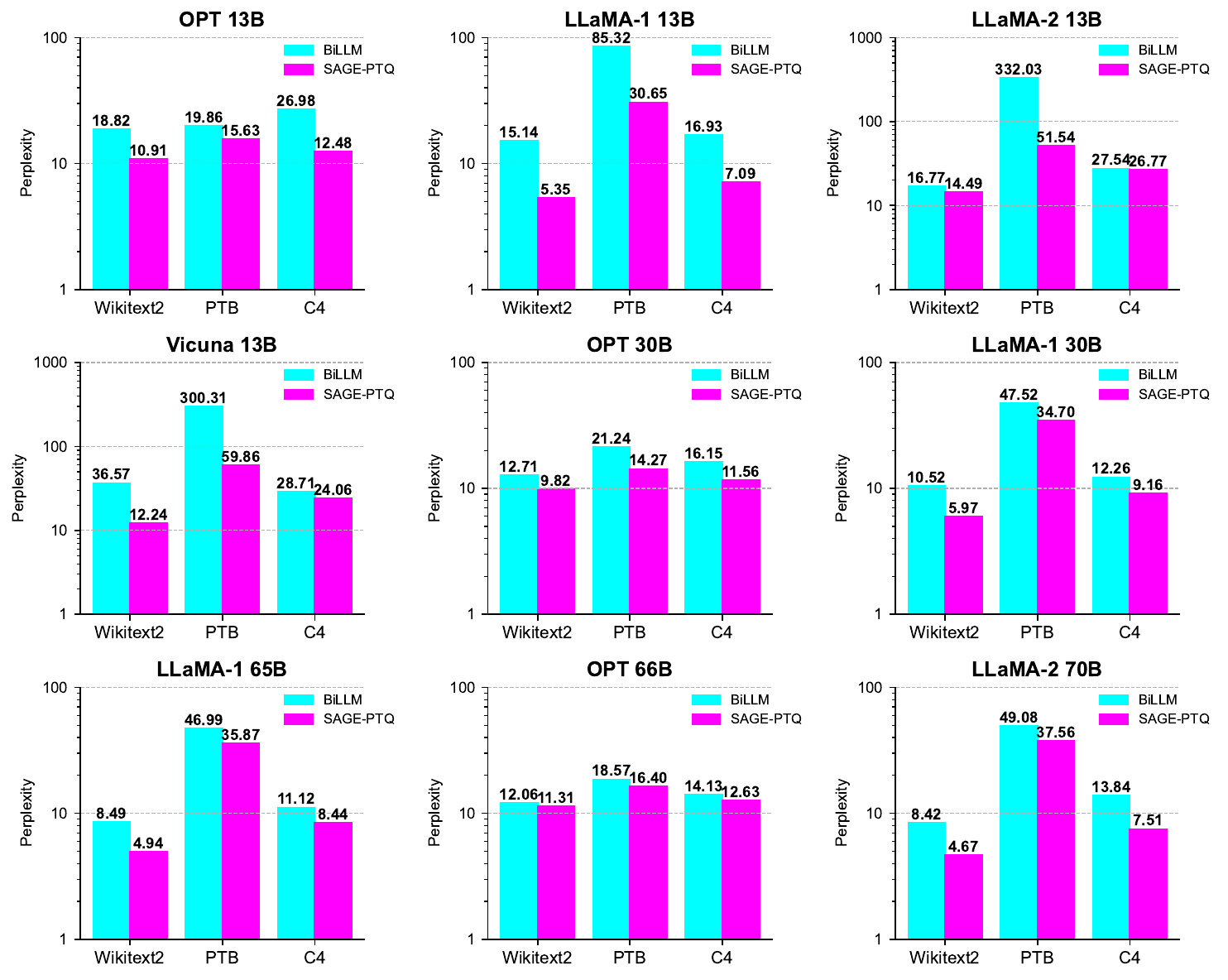}
    \caption{Extended evaluation of \textsc{SAGE-PTQ} versus BiLLM under a lookup table constraint of $N_{\text{LUT}} = 4$ bits across multiple model families (OPT, LLaMA-1, LLaMA-2, and instruction-tuned Vicuna) ranging from 13B to 70B parameters. The bar plots report perplexity on WikiText2, PTB, and C4 datasets. \textsc{SAGE-PTQ} consistently outperforms BiLLM across all models, achieving an average precision of 1.03--1.07 bits per weight.}
    \label{fig:extended}
\end{figure*}
\subsection{Runtime Efficiency}
A key contribution of the \textsc{SAGE-PTQ} method is its ability to reduce scaling overhead during inference to a negligible fraction relative to the size of the weight matrices, while supporting true binary precision. To demonstrate this efficiency, we compare \textsc{SAGE-PTQ} with SoTA ultra-low precision baselines: GPTQ (2-bit), PB-LLM (sub-2-bit), and BiLLM (binary), across three large-scale models—OPT-66B, LLaMA-1 65B, and LLaMA-2 70B. Our evaluation includes all untracked scaling overheads in each method. Results show that \textsc{SAGE-PTQ} achieves the lowest inference-time memory footprint, requiring less than 50\% of the memory used by the most efficient baseline (BiLLM). Moreover, despite operating at binary precision with minimal overhead, our method outperforms all baselines in Wikitext2 perplexity without relying on position-based heuristics, as summarized in Table~\ref{tab4}.
\begin{table*}[!t]
\centering
\caption{Comparison of quantization methods across three large-scale models. SAGE-PTQ achieves the lowest GPU occupancy and perplexity despite using binary weights and negligible scaling overhead.}
\label{tab4}
\begin{tabular}{llcccc}
\toprule
\textbf{Model} & \textbf{Metric} & \textbf{GPTQ} & \textbf{PB-LLM (10\%)} & \textbf{BiLLM} & \textbf{SAGE-PTQ } \\
\midrule

\multirow{5}{*}{OPT 66B}
& Weight Bits          & 2.00   & 1.70   & 1.11  & \textbf{1.07} \\
& Scale Bits           & 0.25   & 0.50   & 1.00  & \textbf{0.009} \\
& Block Size           & 128    & 128    & 128   &\textbf{ N/A} \\
& GPU Occupancy        & 14.06\%& 13.75\%& 13.19\% & \textbf{6.70}\% \\
& Wikitext2 Perplexity & 82.10  & 29.09  & 12.06 & \textbf{11.31 }\\
\midrule

\multirow{5}{*}{LLaMA-1 65B}
& Weight Bits          & 2.00   & 1.70   & 1.09  &\textbf{ 1.03} \\
& Scale Bits           & 0.25   & 0.50   & 1.00  & \textbf{0.004} \\
& Block Size           & 128    & 128    & 128   & \textbf{N/A} \\
& GPU Occupancy        & 14.06\%& 13.75\%& 13.06\% & \textbf{6.4}6\% \\
& Wikitext2 Perplexity & 8.78   & 12.53  & 8.49  & \textbf{4.93} \\
\midrule

\multirow{5}{*}{LLaMA-2 70B}
& Weight Bits          & 2.00   & 1.70   & 1.08  &\textbf{ 1.03} \\
& Scale Bits           & 0.25   & 0.50   & 1.00  & \textbf{0.004} \\
& Block Size           & 128    & 128    & 128   & \textbf{N/A} \\
& GPU Occupancy        & 14.06\%& 13.75\%& 13.00\% & \textbf{6.46\%} \\
& Wikitext2 Perplexity & 9.12   & 28.37  & 8.41  & \textbf{4.67} \\
\bottomrule
\end{tabular}
\end{table*}
\section{Extended Ablation Analysis}
\subsection{Evaluation of Saliecy Metric}
\label{sal_ablation}
\begin{figure*}[!t]
    \centering
    \includegraphics[width=0.9\textwidth,height=8.6cm]{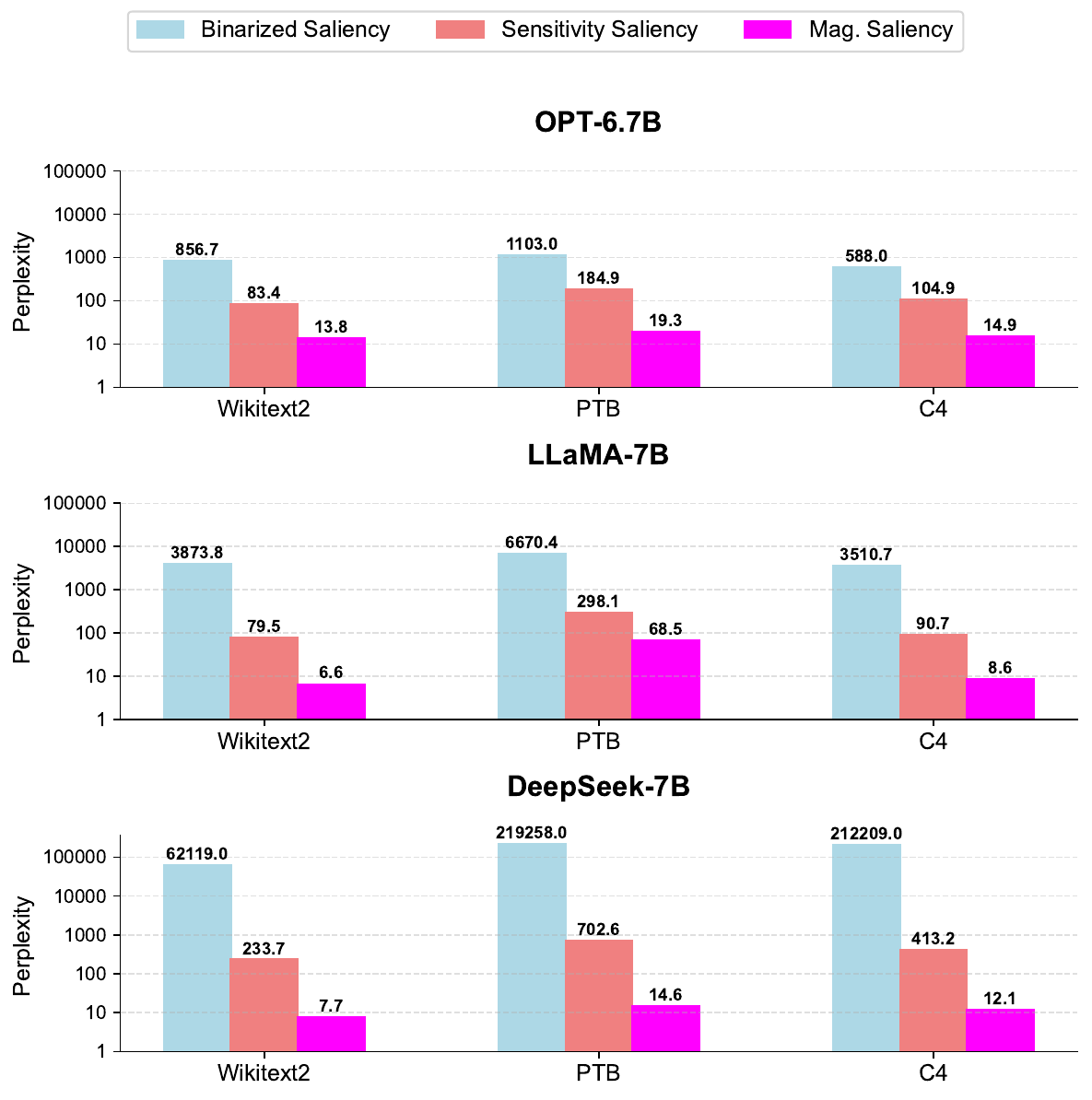}
    \caption{SAGE-PTQ Ablation study on saliency metrics across three model families (OPT-6.7B, LLaMA-7B, DeepSeek-7B) and three datasets (WikiText2, PTB, C4). We compare perplexity results under three conditions: (1) magnitude-based saliency, (2) Hessian-based saliency, and (3) uniform binarization of all weights without saliency distinction. Magnitude-based saliency consistently yields the lowest perplexity, demonstrating its effectiveness and efficiency for SAGE-PTQ method}
    \label{saliency_fig}
    \vspace{-0.1cm}
\end{figure*}

We conducted additional analysis on SAGE-PTQ to compare calibration-based saliency metrics used for identifying quantization-sensitive weights. Specifically, we evaluated the classical Hessian-based sensitivity metric adopted in prior work such as BiLLM, defined as $S_{ij} = \frac{W_{ij}^2}{H_{ii}^{-1}}$, where $H_{ii}^{-1}$ is approximated using Cholesky decomposition of the activation Gram matrix. Our goal was to determine whether second-order statistics are necessary for saliency detection, or if our affinity-based strategy—focused on preserving weight statistics and minimizing quantization error—is more effective. We also evaluated a strategy where all weights, including salient ones, are binarized uniformly to test the importance of preserving high-precision weights. As shown in Figure~\ref{saliency_fig}, the magnitude-based saliency metric outperforms all others across different models and datasets. Uniform binarization of salient weights leads to substantial degradation, confirming their impact on model performance. However, Hessian-based metrics did not yield a significant advantage over simple magnitude, indicating that magnitude alone is a reliable and efficient saliency indicator.

\subsection{Impact of Lookup Table Size on Performance}
\label{lut_ablation}
Quantization resolution plays a critical role in controlling the loss induced by post-training quantization. In \textsc{SAGE-PTQ}, resolution is determined by the number of weight groups, each associated with a unique scaling factor. Increasing the number of groups improves quantization fidelity but requires a larger lookup table (LUT) to store group indices for inference. While \textsc{SAGE-PTQ} ensures that scale overhead remains negligible—requiring only one scalar per group—lookup size introduces a trade-off between performance and idle storage.
\begin{figure*}[!t]
    \centering
    \includegraphics[width=0.80\textwidth,height=8.5cm]{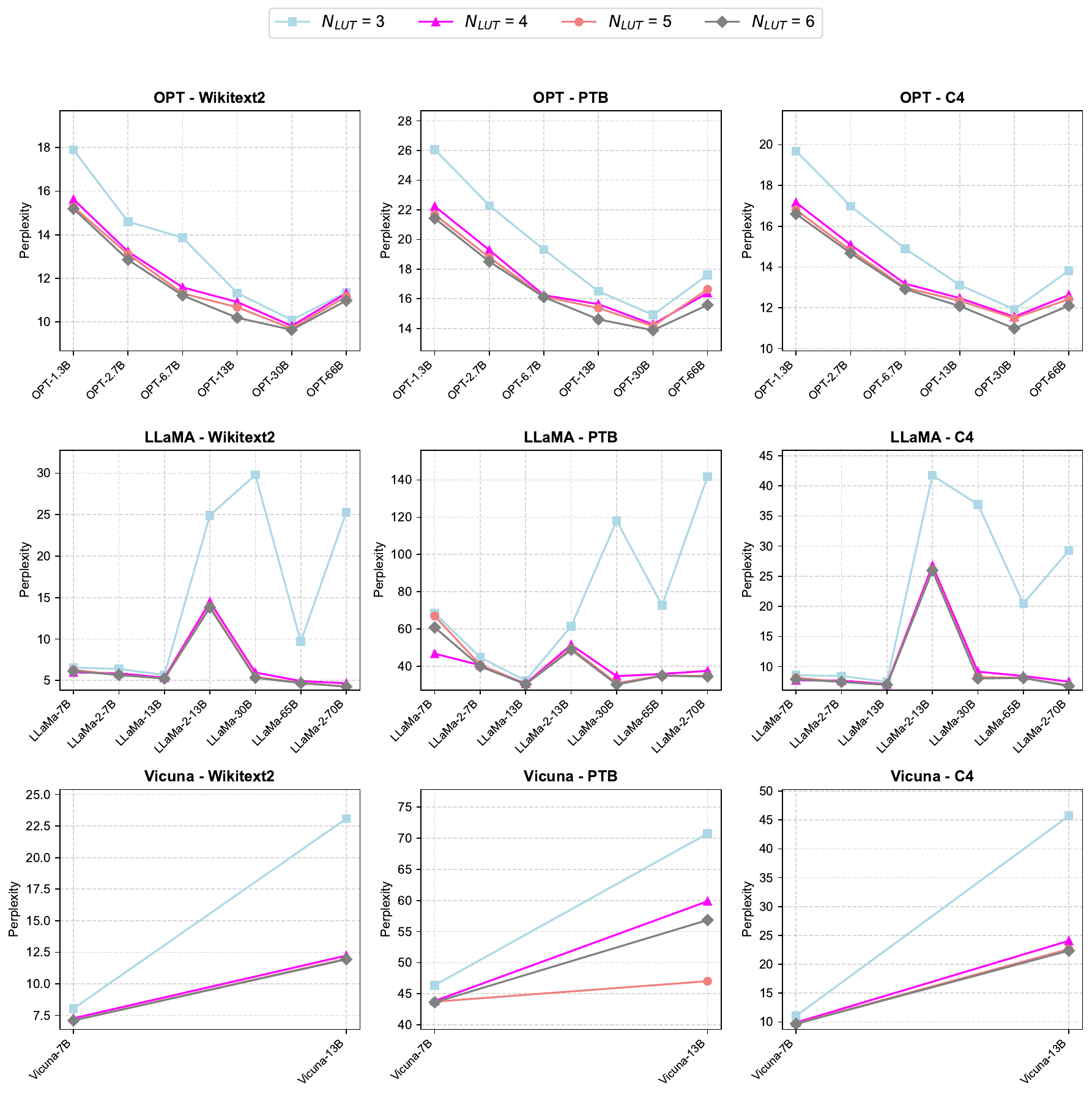}
    \caption{
    Perplexity impact of varying lookup table (LUT) sizes in \textsc{SAGE-PTQ} across OPT (1.3B–66B), LLaMA-1/2 (7B–70B), and Vicuna (7B, 13B) on WikiText2, PTB, and C4. While OPT performs well with 3-bit LUTs, LLaMA and Vicuna benefit from 4+ bits. A 4-bit LUT offers an effective trade-off between accuracy and storage.
    }
    \label{nlut}
    \vspace{-0.5cm}
\end{figure*}
We evaluate the effect of varying lookup sizes (3 to 6 bits, i.e., up to 64 groups) on perplexity across three model families: OPT (1.3B to 66B), LLaMA-1 and LLaMA-2 (7B to 70B), and instruction-tuned Vicuna (7B, 13B). Experiments are conducted on Wikitext2, PTB, and C4 datasets.

As shown in Figure~\ref{nlut}, a 3-bit lookup (8 groups) yields suboptimal performance. However, beyond 4 bits, the perplexity gain saturates. For OPT models, which exhibit stable weight distributions (see Appendix~A), even 3-bit lookups suffice. In contrast, LLaMA models benefit from 4-bit lookups, consistent with their slight deviation from Gaussianity. Instruction-tuned models like Vicuna show notable gains with higher LUT sizes, likely due to fine-tuning introducing specialized weight distributions.
We select a 4-bit LUT as the default setting, offering a good trade-off between accuracy and storage, and enabling fair comparison with prior methods.

\section{Other Design Considerations}\label{hyper_param}

\subsection{Graph Reconstruction Cost}
\paragraph{Graph Construction Cost.}
We employ graph-based modeling to determine the optimal number of unsalient groups for each weight matrix. 
To ensure efficiency, we construct sparse $k$-NN graphs using subsampled unsalient weights, which significantly reduces computational cost while preserving the underlying weight structure. 
All hyperparameters for unsalient weight group estimation are selected via grid search.
After filtering salient weights, we compute summary statistics of the remaining unsalient (inlier) weight distribution and draw a subsample that matches these statistics. 
We evaluate subsampling ratios of \{0.01\%, 0.03\%, 1\%, 3\%\}. 
A ratio of 0.01\% is insufficient to capture the distribution across model sizes, while ratios between 0.03\% and 3\% yield nearly identical grouping results. 
Since larger samples increase runtime without accuracy gains, we select a subsampling ratio of 0.03\%.

For sparse $k$-NN graph construction, we perform a grid search over $k$. 
We observe stable performance at $k=10$ for OPT models and $k=5$ for LLaMA, DeepSeek, and Vicuna. 
This difference is expected, as OPT exhibits higher inlier variance, requiring a larger neighborhood size to capture meaningful weight relations (see Appendix \ref{stat_study}).

\subsection{Method Sensitivity and Hyperparameter Selection.} \label{sensi_param}
We analyze the sensitivity of \textsc{SAGE-PTQ} to key hyperparameters and derive practical default settings.

\textbf{Maximum saliency percentage.}
We analyze weight distributions across multiple model families (Appendix~B) to determine an appropriate upper bound on the saliency percentage. 
We evaluate values in \{0.01, 0.05, 0.1, 0.5\}. 
Adaptive saliency thresholding consistently selects saliency ratios below 0.1 for OPT models and below 0.01 for LLaMA, DeepSeek, and Vicuna. 
This behavior aligns with the underlying statistics: OPT exhibits shorter tails with higher variance, while LLaMA, DeepSeek, and Vicuna have longer tails with lower variance but larger outliers. 
We quantify these differences using KL divergence (Appendix~A, Figures~7--9).

\textbf{Outlier bit budget.}
Performance improves monotonically with increased outlier precision. 
Based on this trend, we establish minimal precision settings that preserve accuracy. 
We use a 0.01 saliency ratio with a 4-bit outlier budget for long-tail models (LLaMA 7B--70B, DeepSeek 7B, Vicuna 7B/13B), and a 0.1 saliency ratio with 2-bit precision for short-tail models (OPT 1.3B--66B).

\textbf{Lookup size.}
We evaluate lookup bit budgets \( N_{\text{LUT}} \in \{2,3,4,5,6\} \), where each budget supports up to \( 2^{N_{\text{LUT}}} \) weight groups. 
Across all model families, performance stabilizes for \( N_{\text{LUT}} > 2 \) and remains unchanged beyond \( N_{\text{LUT}} = 3 \). 
We therefore adopt \( N_{\text{LUT}} = 4 \) as a robust default. 
Additional analysis is provided in and Appendix \ref{lut_ablation}.

Overall, these settings represent minimal configurations; larger budgets further improve performance while preserving robustness.



\end{document}